\documentclass[12pt]{article}

\usepackage{mathtools}
\usepackage{booktabs}
\usepackage[english]{babel} 
\usepackage[protrusion=true,expansion=true]{microtype} 
\usepackage{amsmath,amsfonts,amsthm}
\usepackage{ amssymb }
\usepackage{enumitem}
\usepackage{graphicx}
\usepackage{array}
\usepackage[toc,page]{appendix}
\usepackage{xcolor}
\usepackage{tikz}
\usetikzlibrary{arrows.meta,positioning}

\usepackage{booktabs} 
\usepackage{natbib}
\usepackage{setspace}
\usepackage{microtype} 
\usepackage{tabularx}
\usepackage{subcaption}
\usepackage[font = small,labelfont=bf,textfont=it]{caption} 
\usepackage{footnote}
\usepackage{algorithm}
\usepackage[noend]{algpseudocode}
\usepackage{multirow}
\usepackage{url}

\makeatletter
\newcommand{\distas}[1]{\mathbin{\overset{#1}{\kern\z@\sim}}}%
\newsavebox{\mybox}\newsavebox{\mysim}

\theoremstyle{definition}

\newcommand{\distras}[1]{%
  \savebox{\mybox}{\hbox{\kern3pt$\scriptstyle#1$\kern3pt}}%
  \savebox{\mysim}{\hbox{$\sim$}}%
  \mathbin{\overset{#1}{\kern\z@\resizebox{\wd\mybox}{\ht\mysim}{$\sim$}}}%
}
\newcolumntype{C}[1]{>{\centering\let\newline\\\arraybackslash\hspace{0pt}}m{#1}}

\newcolumntype{Y}{>{\centering\arraybackslash}X}

\newcommand{\blind}{1}
\newcommand{\bl}{\textcolor{blue}}

\begin{document}

\def\spacingset#1{\renewcommand{\baselinestretch}%
{#1}\small\normalsize} \spacingset{1.3}


\if1\blind
{
 \centering{\bf\Large 
 Deep Intrinsic Coregionalization Multi-Output Gaussian Process Surrogate with Active Learning}\\
  \vspace{0.3in}
  \centering{Chun-Yi Chang and Chih-Li Sung
  \vspace{0.1in}\\
        Michigan State University\\
        }
    \date{\vspace{-7ex}}
} \fi

\if0\blind
{
  \bigskip
  \bigskip
  \bigskip
    \begin{center}
    {\LARGE\bf Deep Intrinsic Coregionalization Multi-Output Gaussian Process Surrogate with Active Learning}
\end{center}
  \medskip
} \fi

\bigskip
\begin{abstract}
    Deep Gaussian Processes (DGPs) are powerful surrogate models known for their flexibility and ability to capture complex functions. However, extending them to multi-output settings remains challenging due to the need for efficient dependency modeling. We propose the Deep Intrinsic Coregionalization Multi-Output Gaussian Process (deepICMGP) surrogate for  computer simulation experiments involving multiple outputs, which extends the Intrinsic Coregionalization Model (ICM) by introducing hierarchical coregionalization structures across layers. This enables deepICMGP to effectively model nonlinear and structured dependencies between multiple outputs, addressing key limitations of traditional multi-output GPs. We benchmark deepICMGP against state-of-the-art models, demonstrating its competitive performance. Furthermore, we incorporate active learning strategies into deepICMGP to optimize sequential design tasks, enhancing its ability to efficiently select informative input locations for multi-output systems.
\end{abstract}

\noindent%
\vfill

\newpage
\spacingset{1.45} 

\section{Introduction}
Multi-output surrogate modeling is crucial for computer simulation experiments, particularly in computationally intensive domains such as aerospace engineering and climate science, where multiple responses must be predicted efficiently. In many real-world applications, including dynamic systems \citep{ARIZARAMIREZ201826}, airfoil aerodynamics \citep{liu2014modeling, lin2022gradient}, and catalytic oxidation processes \citep{wang2015gaussian}, multi-output simulations commonly arise and are computational demanding. Effectively emulating these simulations requires surrogate models that can capture dependencies among outputs while maintaining both predictive accuracy and computational efficiency.

Among various surrogate modeling techniques, Gaussian processes (GPs) \citep{rasmussen2005gaussian, gramacy2020surrogates}, also known as \textit{kriging}, are widely used due to their flexibility and ability to quantify predictive uncertainty. For problems involving multiple correlated outputs, Multi-Output Gaussian Processes (MOGPs) extend the standard GP framework to jointly model these outputs while capturing their interdependencies. A classical example is co-kriging \citep{matheron1969universel, myers1982matrix}, also known as multivariate kriging, which improves predictive performance by incorporating information from correlated variables within a unified model.

Several MOGP frameworks have been developed to explicitly model output correlations. One of the most widely used is the linear model of coregionalization (LMC) \citep{goovaerts1997geostatistics}, which represents each output as a linear combination of shared latent functions. A special case of LMC is the intrinsic coregionalization model (ICM) \citep{NIPS2007_bonilla}, in which all outputs share the same latent structure. The ICM has been successfully applied in diverse areas such as computer model emulation \citep{CONTI2010640} and classification \citep{skolidis2011bayesian}. Another notable extension is the Semiparametric Latent Factor Model (SLFM) \citep{Teh2005latentfactor}, which enriches the LMC framework by introducing structured latent variables. In addition, the convolved Gaussian process model \citep{alvarez2008, JMLR:v12:alvarez11a} constructs each output by convolving latent GPs with smoothing kernels, thereby inducing dependencies across output variables. Other approaches include spectral methods such as the Spectral Mixture Linear Model of Coregionalization (SM-LMC) \citep{wilson2014covariance}, the Cross Spectral Mixture (CSM) kernel \citep{CSM}, and the Multi-Output Spectral Mixture (MOSM) kernel \citep{MOSM}, which model output dependencies in the frequency domain.

In this article, we propose a new MOGP built upon the deep Gaussian process (DGP) framework developed by \citet{Sauer02012023}, utilizing posterior inference via a hybrid Markov Chain Monte Carlo (MCMC) scheme. A DGP composes multiple GP layers to form a hierarchical model, originally introduced by \citet{pmlr-v31-damianou13a} using variational inference. However, variational methods often struggle to adequately quantify uncertainty. To address this, \citet{Sauer02012023} proposed using MCMC-based techniques to better capture posterior uncertainty. While DGPs are flexible and well-suited for modeling complex, non-stationary functions, they are originally designed for scalar outputs, limiting their direct applicability to multi-output problems. The non-stationary nature of DGPs, however, makes them particularly attractive for sequential design tasks, motivating our adoption of this framework.

In particular, we incorporate ICM structures into the DGP to model dependencies across multiple outputs via a shared covariance structure, yielding a Deep Intrinsic Coregionalization Multi-Output Gaussian Process (deepICMGP). The ICM-based covariance structure not only better captures the correlations among highly related outputs but also simplifies the training of the DGP. Furthermore, the non-stationary nature of DGPs enables effective modeling of non-stationary multi-output data.

Building on the newly deepICMGP surrogate, we propose an active learning framework for multi-output computer experiments to enhance predictive accuracy while efficiently managing limited computational resources. In contrast to space-filling or one-shot designs, such as maximin Latin hypercube designs \citep{MORRIS1995381MmLHD} and maximum projection designs \citep{joseph2015maximum},  active learning (AL), also known as \textit{sequential designs}, provides a more efficient alternative by iteratively selecting the most informative input points to maximize information gain relative to computational cost \citep{gramacy2020surrogates}. Guided by this principle, we develop an active learning strategy by tailoring the Active Learning Cohn (ALC) criterion \citep{Cohn1993} to our proposed deepICMGP surrogate, allowing for efficient and targeted selection of input locations that benefit all outputs simultaneously. This approach is especially well-suited to scenarios with constrained evaluation budgets, where maximizing information gain across correlated outputs is essential for improving overall predictive performance.


Within this framework, our main contributions are as follows: (i) we propose a deepICMGP surrogate for multi-output computer simulations---a novel multi-output DGP that integrates the ICM to capture complex inter-output correlations through nonlinear warping;
(ii) we demonstrate computational efficiency by leveraging shared latent representations and covariance structures during training; and
(iii) we introduce a simultaneous sequential design strategy by adapting the ALC criterion to select a single informative input location for all outputs based solely on latent representations. 

The remainder of this paper is organized as follows. Section~\ref{sec:existing MOGP} reviews existing MOGP approaches. Section~\ref{sec:deepICM} introduces the proposed deepICMGP surrogate. In Section~\ref{sec:AL}, we present a tailored ALC strategy for efficient sequential design using the  deepICMGP surrogate. Section~\ref{sec:numerical} reports extensive numerical studies benchmarking our method against existing MOGP approaches and demonstrating the performance of the ALC framework under the deepICMGP surrogate. We further illustrate the effectiveness of deepICMGP through a real-world case study on thermal stress analysis in Section~\ref{sec:real_data_analysis}. Finally, Section~\ref{sec:conclusion} concludes with remarks and future directions.



\section{Background}\label{sec:existing MOGP}


\subsection{Problem Setup}
Consider a black-box function that simultaneously outputs $Q$ responses, $\mathbf{f}:=(f_1,\ldots,f_Q): \mathbb{R}^d \rightarrow \mathbb{R}^Q$, which represent an expensive computer model or simulator involving $Q$ outputs in the context of computer experiments. A surrogate model is developed to approximate $\mathbf{f}$ based on $n$ selected training design points. Let $\mathbf{X}_n=(\mathbf{x}_1,\mathbf{x}_2,\dots, \mathbf{x}_n)^\top \in \mathbb{R}^{n \times d}$ denote the design matrix, where each row $\mathbf{x}_i \in \mathbb{R}^d$ represents a $d$-dimensional input location. In this work, we focus on \textit{deterministic} computer models, i.e., simulations without observation noise. Accordingly, the output vector for the $q$-th output is defined as $\mathbf{y}_q=f_q(\mathbf{X}_n)=\left[f_q(\mathbf{x}_1),\dots, f_q(\mathbf{x}_n)\right]^\top\in \mathbb{R}^n$. The full output matrix is given by $\mathbf{Y}_n=(\mathbf{y}_1, \mathbf{y}_2, \dots, \mathbf{y}_Q)\in \mathbb{R}^{n \times Q}$, and the vectorized output is $\mathbf{y}_{nQ}=\mathrm{vec}({\mathbf{Y}_n})\in \mathbb{R}^{nQ}$, where $\mathrm{vec}(\cdot)$ denotes the column-wise vectorization operator that stacks the columns of a matrix into a single vector.

\subsection{Intrinsic Coregionalization Model (ICM)}\label{sec:ICM}
A Gaussian process (GP) is a popular nonparametric tool for modeling nonlinear regression problems \citep{rasmussen2005gaussian,gramacy2020surrogates}. Specifically, a GP assumes that the response is a realization of a stochastic process, such that any finite collection of responses follows a multivariate normal distribution. However, conventional GPs are typically limited to scalar outputs. To extend GPs to multi-output settings, one of the most commonly used approaches is the intrinsic coregionalization model (ICM), which expresses each output as a linear combination of independent latent functions, each following a GP model. The ICM serves as a foundational framework for modeling output dependencies and was employed by \citet{NIPS2007_bonilla} for multi-task learning. It captures correlations among outputs by learning a shared covariance structure through input-dependent features.

Specifically, the ICM models each output as a linear combination of $R$ shared latent functions:
\begin{equation}\label{eq:ICM}
    f_q(\mathbf{x}) = \sum_{r=1}^R a^r_q\, u^r(\mathbf{x}),\quad\text{for}\quad q=1,\ldots,Q,
\end{equation}
where each latent function follows a GP with zero mean and a common positive-definite kernel $K_{\theta}(\mathbf{x}, \mathbf{x}')$ with the hyperparameter $\theta$,  i.e., $u^r(\mathbf{x}) \sim \mathcal{GP}(0, K_{\theta}(\mathbf{x}, \mathbf{x}'))$ for $r=1,\ldots,R$. Then, the output covariance at inputs $\mathbf{x}$ and $\mathbf{x}'$ becomes
$$\mathrm{Cov}(\mathbf{f}(\mathbf{x}), \mathbf{f}(\mathbf{x}')) = \mathbf{A} \mathbf{A}^\top K_{\theta}(\mathbf{x}, \mathbf{x}') = \mathbf{B} \, K_{\theta}(\mathbf{x}, \mathbf{x}'),$$
where $ \mathbf{A} = [\mathbf{a}^1, \dots, \mathbf{a}^R] \in \mathbb{R}^{Q \times R} $, $\mathbf{a}^r=(a^r_1,\ldots,a^r_q)^T$, $ \mathbf{B} = \mathbf{A} \mathbf{A}^\top \in \mathbb{R}^{Q \times Q} $, and $ \mathrm{rank}(\mathbf{B}) \leq R $. 

Given the training inputs $\mathbf{X}_n$, the model \eqref{eq:ICM} implies that the corresponding vectorized outputs $\mathbf{y}_{nQ}$ follow an $nQ$-dimensional multivariate normal distribution:
$$\mathbf{y}_{nQ}\sim \mathcal{N}_{nQ}(\mathbf{0}, \mathbf{B} \otimes K_{\theta}(\mathbf{X}_n)),$$
where $\otimes$ denotes the Kronecker product, and $K_{\theta}(\mathbf{X}_n)$ is an $n\times n$ matrix with each element $[K_{\theta}(\mathbf{X}_n)]_{i,j}=K_{\theta}(\mathbf{x}_i, \mathbf{x}_j)$. Note that if the data are noisy instead of deterministic, a nugget term can be included in $K_{\theta}(\mathbf{X}_n)$, yielding $[K_{\theta}(\mathbf{X}_n)]_{i,j}=K_{\theta}(\mathbf{x}_i, \mathbf{x}_j)+g\delta(\mathbf{x}_i, \mathbf{x}_j)$, where $g > 0$ is the nugget parameter and $\delta(\mathbf{x}_i, \mathbf{x}_j)$ is the Kronecker delta, equal to 1 when $\mathbf{x}_i=\mathbf{x}_j$ and 0 otherwise. In this formulation, $\mathbf{B}$ captures the covariance across outputs, while $K_{\theta}(\mathbf{X}_n)$ captures dependencies across input locations. The correlation form of $\mathbf{B}$ is also referred to as the \textit{cross-correlation} matrix \citep{qian2008gaussian, zhou2011simple}. This Kronecker-separable  structure characterizes the ICM as a separable kernel model,  offering computational efficiency and interpretability in multi-output settings. This approach is also known as  the Separable Linear Model of Coregionalization (SLMC)  \citep{CONTI2010640} and has been widely used in various applications \citep{separable2007bhattacharya, separable2008kennedy}.

Based on the properties of conditional normal distributions, it can be shown that the predictive distribution of $\mathbf{f}(\mathbf{x}')$ at any new input $\mathbf{x}'$, given the training data, follows a multivariate normal distribution (details provided in Section~\ref{sec:prediction}). Interestingly, in the deterministic setting, the predictive mean under ICM becomes equivalent to that of an independent GP model with a diagonal $\mathbf{B}$ matrix \citep{helterbrand1994universal, mak2018efficient}. This implies that the coregionalization matrix $\mathbf{B}$ has no effect on the predictive mean. Our proposed approach, developed in Section~\ref{sec:deepICM}, overcomes this limitation by forming a hierarchical model using a deep GP, which enables richer modeling of output dependencies even in deterministic scenarios. 


A more general framework than ICM is the Linear Model of Coregionalization (LMC) \citep{journel1978mining, goovaerts1997geostatistics, zhang2007maximum,Fricker01022013, wackernagel2013multivariate}, which allows each latent function $u^r(\mathbf{x}) \sim \mathcal{GP}(0, K_{\theta_r}(\mathbf{x}, \mathbf{x}'))$ to have its own distinct covariance kernel $K_{\theta_r}$. While this additional flexibility enables LMC to capture more complex dependencies across outputs,  it also increases computational complexity for inference as it requires to impose contraints to ensure positive definiteness of the cross-output covariance in the inference. In contrast, the Kronecker product structure of ICM enables more efficient inference, particularly in large-scale problems, by allowing the full covariance matrix $\mathbf{B} \otimes \mathbf{K}_n$ to be inverted more efficiently. 

\section{Deep Intrinsic Coregionalization Model}\label{sec:deepICM}



To leverage both the computational efficiency and output-dependence modeling of ICM, along with the flexibility of deep Gaussian processes (DGPs) in capturing complex, non-stationary relationships, we propose a new model: the Deep Intrinsic Coregionalization Multi-Output Gaussian Process (deepICMGP). Specifically, the deepICMGP surrogate integrates the ICM framework into the layered structure of a DGP, built upon the hybrid MCMC inference framework developed by \citet{Sauer02012023}, to enable flexible and scalable modeling of correlated multi-output systems.

Here, we focus on a two-layer deepICMGP model, though it can be readily extended to more than two layers. The deepICMGP introduces a latent layer  $\mathbf{W}_n$, which serves as a transformation or ``warping'' of the input matrix $\mathbf{X}_n$. Specifically, we define $\mathbf{W}_n=(W_1, W_2, \dots, W_D)\in \mathbb{R}^{n \times D}$, where $D$ is the number of latent nodes. The corresponding vectorized form is $\mathbf{w}_{nD}=\mathrm{vec}{(\mathbf{W}_n})\in \mathbb{R}^{nD}$. 

We assume that the latent layer $\mathbf{W}_n$ follows an ICM model based on the inputs $\mathbf{X}_n$, denoted as $\mathbf{W}_n=\mathbf{f}_1(\mathbf{X}_n)$, and that the final output layer $\mathbf{Y}_n$ also follows an ICM model, now conditioned on $\mathbf{W}_n$, denoted as $\mathbf{Y}_n=\mathbf{f}_2(\mathbf{W}_n)$. This results in the following hierarchical structure:
\begin{align}
    \mathbf{y}_{nQ}\mid \mathbf{W}_n, \mathbf{B}_y, \theta_y \sim \mathcal{N}_{nQ}\Big(\mathbf{0}, \mathbf{B}_y\otimes K_{\theta_y}(\mathbf{W}_n)\Big),\label{eq:Y_dist}\\
   \mathbf{w}_{nD}\mid \mathbf{X}_n, \mathbf{B}_w, \theta_w \sim \mathcal{N}_{nD}\Big(\mathbf{0}, \mathbf{B}_w \otimes K_{\theta_w}(\mathbf{X}_n)\Big),\label{eq:W_dist}
\end{align}
where $\mathbf{B}_y \in \mathbb{R}^{Q\times Q}$ and $\mathbf{B}_w \in \mathbb{R}^{D\times D}$ are the inner and outer coregionalization matrices, respectively. The matrices $K_{\theta_y}(\mathbf{W}_n)$ and $K_{\theta_w}(\mathbf{X}_n)$ are both  $n\times n$ kernel matrices, with entries defined as $[K_{\theta_y}(\mathbf{W}_n)]_{i,j}=K_{\theta_y}(\mathbf{w}_i, \mathbf{w}_j)$ and $[K_{\theta_w}(\mathbf{X}_n)]_{i,j}=K_{\theta_w}(\mathbf{x}_i, \mathbf{x}_j)$, respectively. To ensure numerical stability, a small \textit{nugget} term, such as \texttt{sqrt(.Machine\$double.eps)} in \textsf{R}, which is approximately $10^{-8}$, is typically added to the diagonal of each kernel matrix \citep{gramacy2012cases}.

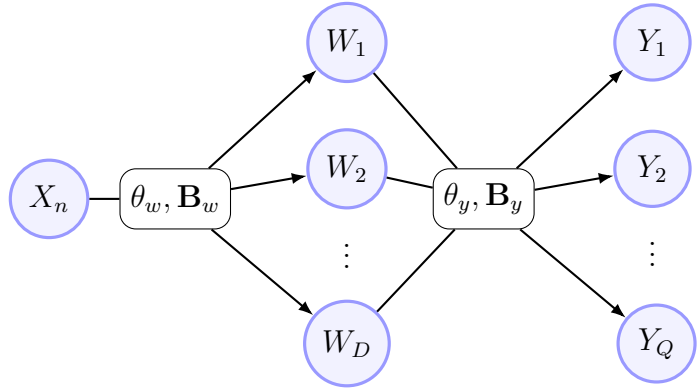
\begin{figure}[t]
    \centering
    \scalebox{1}{
    \begin{tikzpicture}[
    node distance=3cm and 3cm,
    roundnode/.style={circle, draw=blue!40, fill=blue!5, very thick, minimum size=10mm},
    annotation/.style={draw=none, font=\small},
    arrow/.style={-{Latex[length=2mm]}, thick},
    line/.style={-, thick},
    roundedbox/.style={
    draw,
    rectangle,
    rounded corners=6pt,
    minimum height=8mm,
    minimum width=12mm,
    align=center}
]
    \node[roundnode](X){$X_n$};
    \node[roundedbox](paraW)[right=0.4cm of X]{$\theta_w, \mathbf{B}_w$};
    \node[roundnode](W2)[right=1cm of paraW, yshift=0.4cm]{$W_2$};
    \node[roundnode](W1)[above=0.6cm of W2]{$W_1$};
    \node[roundnode](WD)[below=1.2cm of W2]{$W_D$};
    
    \node[roundedbox](paraY)[right=0.6cm of W2, yshift=-0.4cm]{$\theta_y, \mathbf{B}_y$};
    \node[roundnode](Y1)[right=3cm of W1]{$Y_1$};
    \node[roundnode](Y2)[right=3cm of W2]{$Y_2$};
    \node[roundnode](YQ)[right=3cm of WD]{$Y_Q$};
    
    \node[draw=none, below=0.1cm of W2] (dots1) {$\vdots$};
    \node[draw=none, below=0.1cm of Y2] (dots2) {$\vdots$};
    
    \draw[line] (X) -- (paraW);
    
    \draw[arrow] (paraW) -- (W1);
    \draw[arrow] (paraW) -- (W2);
    \draw[arrow] (paraW) -- (WD);
    
    \draw[line] (W1) -- (paraY);
    \draw[line] (W2) -- (paraY);
    \draw[line] (WD) -- (paraY);
    
    \draw[arrow] (paraY) -- (Y1);
    \draw[arrow] (paraY) -- (Y2);
    \draw[arrow] (paraY) -- (YQ);
    \end{tikzpicture}
    }
    \caption{Model framework of a two-layer deepICMGP.}
    \label{fig:deepICMGP}
\end{figure}

Figure \ref{fig:deepICMGP}  illustrates the two-layer structure of the deepICMGP model. Conceptually, the model represents a composite function $\mathbf{Y}_n=\mathbf{f}_2(\mathbf{f}_1(\mathbf{X}_n))$, where both $\mathbf{f}_1$ and $\mathbf{f}_2$ are modeled using ICM structures. The first layer is governed by parameters $(\mathbf{B}_w, \theta_w)$, and the second by $(\mathbf{B}_y, \theta_y)$. This hierarchical formulation can be naturally extended to deeper architectures by stacking additional ICM layers. It is worth noting that our model is intrinsically different from the LMC, another extension of ICM, which uses a linear combination of multiple kernels with a coregionalization matrix. In contrast, our model introduces a nonlinear warping through a hierarchical composition of ICM layers, providing greater flexibility in modeling output relationships. Such deep structures have been shown to be particularly effective for capturing complex and non-stationary functions \citep{Sauer02012023,Ming03042023}.

In this paper, we adopt an isotropic squared-exponential kernel defined as
\begin{align}\label{eq:sq_exp}
    K_{\theta}(\mathbf{x}_i, \mathbf{x}_j)=\exp\left(-\frac{\|\mathbf{x}_i-\mathbf{x}_j\|^2}{\theta}\right), 
\end{align}
where $\theta$ denotes the lengthscale parameter. While this kernel assumes isotropy, the incorporation of coregionalization matrices introduces anisotropic behavior across latent processes and outputs. This effectively allows the model to adjust its response to different input directions through shared latent representations. For the number of latent nodes $D$, we find empirically that setting it to the maximum of the input and output dimensions, i.e., $D=\max(d, Q)$, yields good performance across a variety of settings..

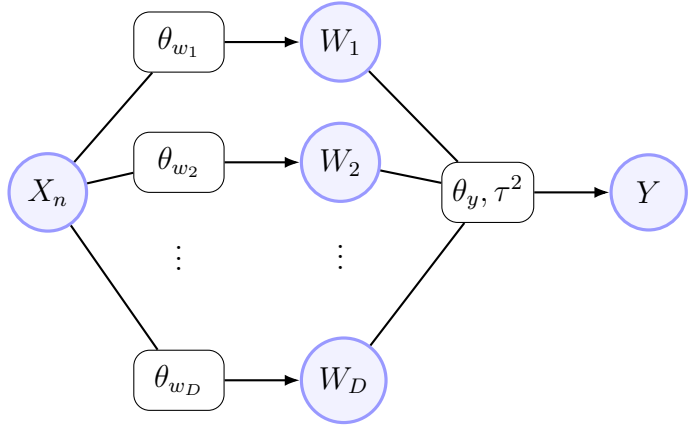
\begin{figure}[t]
    \centering
    \begin{tikzpicture}[
    node distance=3cm and 3cm,
    roundnode/.style={circle, draw=blue!40, fill=blue!5, very thick, minimum size=10mm},
    annotation/.style={draw=none, font=\small},
    arrow/.style={-{Latex[length=2mm]}, thick},
    line/.style={-, thick},
    roundedbox/.style={
    draw,
    rectangle,
    rounded corners=6pt,
    minimum height=8mm,
    minimum width=12mm,
    align=center}
]
    \node[roundnode](X){$X_n$};
    \node[roundedbox](paraW1)[right=0.6cm of X, yshift=2cm]{$\theta_{w_1}$};
    \node[roundedbox](paraW2)[right=0.6cm of X, yshift=0.4cm]{$\theta_{w_2}$};
    \node[roundedbox](paraWD)[right=0.6cm of X, yshift=-2.5cm]{$\theta_{w_D}$};

    \node[roundnode](W1)[right=1cm of paraW1]{$W_1$};
    \node[roundnode](W2)[right=1cm of paraW2]{$W_2$};
    \node[roundnode](WD)[right=1cm of paraWD]{$W_D$};

    \node[roundedbox](paraY)[right=0.8cm of W2, yshift=-0.4cm]{$\theta_{y}, \tau^2$};

    \node[roundnode](Y)[right=1cm of paraY]{$Y$};

    \node[draw=none, below=0.2cm of W2] (dots1) {$\vdots$};
    \node[draw=none, below=0.35cm of paraW2] (dots4) {$\vdots$};

    \draw[line] (X) -- (paraW1);
    \draw[line] (X) -- (paraW2);
    \draw[line] (X) -- (paraWD);

    \draw[arrow] (paraW1) -- (W1);
    \draw[arrow] (paraW2) -- (W2);
    \draw[arrow] (paraWD) -- (WD);

    \draw[line] (W1) -- (paraY);
    \draw[line] (W2) -- (paraY);
    \draw[line] (WD) -- (paraY);

    \draw[arrow] (paraY) -- (Y);
    \end{tikzpicture}
    \caption{Model framework of a two-layer DGP introduced in \cite{Sauer02012023}. }
    \label{fig:deepGPmulti}
\end{figure}

A key distinction between deepICMGP and conventional DGPs lies in the modeling of latent layers. Most existing DGPs are designed for scalar outputs and assume independence across latent dimensions (see Figure \ref{fig:deepGPmulti}). In contrast,  deepICMGP introduces structured dependencies using coregionalization matrices at each layer. This enables it to capture both nonlinear transformations and cross-dimensional correlations, making it well-suited for complex multi-output relationships.

\subsection{Priors}\label{sec:priors}

We adopt a Bayesian approach for inference in the proposed deepICMGP model, with prior specifications detailed in this section. To ensure the kernel parameters $(0, \infty)$, we follow the approach of \citet{Sauer02012023} and assign Gamma priors:
\begin{equation}\label{eq:priorlengthscale}
\{\theta_y, \theta_w\} \overset{\mathrm{i.i.d}}{\sim} \mathrm{Gamma}\left(\frac{3}{2}, b_{\left[\cdot\right]}\right),    
\end{equation}
where the shape parameter is $\frac{3}{2}$ and the rate parameters are set as $b_{\left[ \theta_y \right]}=\frac{3.9}{6}$ and $b_{\left[ \theta_w \right]}=\frac{3.9}{4}$. . While \citet{Sauer02012023} recommend setting $b_{\left[ \theta_y \right]} >b_{\left[ \theta_w \right]}$ to encourage smoother behavior in deeper layers, our choice yields a higher prior mean for $\theta_w$ than for $\theta_y$. This aligns with our intuition that the latent layer should be smoother, capturing broad, abstract transformations that serve as a shared foundation, while the output layer focuses on finer-scale variations tailored to each response.


We adopt Jeffreys-type non-informative priors for the coregionalization matrices $\mathbf{B}_w$ and $\mathbf{B}_y$, following \citet{CONTI2010640}:
$$\pi(\mathbf{B}_w) \propto |\mathbf{B}_w|^{-\frac{(D+1)}{2}},\quad\text{and}\quad\pi(\mathbf{B}_y) \propto |\mathbf{B}_y|^{-\frac{(Q+1)}{2}}.$$
We assume $\mathbf{B}_w$ and $\mathbf{B}_y$ are independent of each other, and also independent of the lengthscale parameters $\theta_w$ and $\theta_y$.


\subsection{MCMC Posterior Sampling}\label{sec:posterior}
Following \citet{Sauer02012023}, we employ a hybrid Gibbs–Metropolis–Hastings (MH) and elliptical slice sampling (ESS) scheme \citep{pmlr-v9-murray10a}. One of the major challenges in ICM and its extension, LMC, lies in estimating the coregionalization matrices, which involve many unknown parameters and must be constrained to remain positive semi-definite \citep{qian2008gaussian}. To address this,  we integrate out the coregionalization matrices $\mathbf{B}_w$ and $\mathbf{B}_y$ under the non-informative priors given in Section~\ref{sec:priors}, yielding a closed-form marginal likelihood that enhances computational efficiency. Posterior inference is then carried out for the hyperparameters ${\theta_y, \theta_w}$ and the latent layer $\mathbf{W}_n$. Specifically, scalar hyperparameters are updated via MH with adaptive proposals, while the latent variables $\mathbf{W}_n$ are sampled using ESS. The Gibbs sampler coordinates the procedure by iteratively updating each component conditional on the others.

Based on \eqref{eq:Y_dist} and \eqref{eq:W_dist}, the likelihood is given by 
\begin{align}
    \mathcal{L}(\mathbf{Y}_n\mid \mathbf{W}_n, \mathbf{B}_y, \theta_y) &\propto  |K_{\theta_y}(\mathbf{W}_n)|^{-\frac{Q}{2}} |\mathbf{B}_y|^{-\frac{n}{2}} \exp\Big\{-\frac{1}{2} \operatorname{tr}\left(\mathbf{Y}_n^\top K_{\theta_y}(\mathbf{W}_n)^{-1}\mathbf{Y}_n\mathbf{B}_y^{-1}\right)\Big\},\nonumber\\
    \mathcal{L}(\mathbf{W}_n\mid \mathbf{X}_n, \mathbf{B}_w, \theta_w) &\propto  |K_{\theta_w}(\mathbf{X}_n)|^{-\frac{D}{2}} |\mathbf{B}_w|^{-\frac{n}{2}} \exp\Big\{-\frac{1}{2} \operatorname{tr}\left(\mathbf{W}_n^\top K_{\theta_w}(\mathbf{W}_n)^{-1}\mathbf{W}_n\mathbf{B}_w^{-1}\right)\Big\}.\label{eq:marglikW}
\end{align}
Integrating out $\mathbf{B}_y$ and $\mathbf{B}_w$ using the Jeffreys-type priors yields the marginal likelihoods:
\begin{align*}
    \mathcal{L}(\mathbf{Y}_n\mid \mathbf{W}_n,\theta_y) &\propto |K_{\theta_y}(\mathbf{W}_n)|^{-\frac{Q}{2}} |n\hat{\mathbf{B}}_y|^{-\frac{n}{2}} ,\\
    \mathcal{L}(\mathbf{W}_n\mid \mathbf{X}_n, \theta_w) &\propto   |K_{\theta_w}(\mathbf{X}_n)|^{-\frac{D}{2}} |n\hat{\mathbf{B}}_w|^{-\frac{n}{2}},
\end{align*}
where $$\hat{\mathbf{B}}_w=\frac{\mathbf{W}^\top _nK_{\theta_w}(\mathbf{X}_n)\mathbf{W}_n}{n} \quad\text{and}\quad \hat{\mathbf{B}}_y=\frac{\mathbf{Y}^\top _nK_{\theta_y}(\mathbf{W}_n)\mathbf{Y}_n}{n},$$ which can be seen as a generalized least squares (GLS) estimator of $\mathbf{B}_w$ and $\mathbf{B}_y$ \citep{CONTI2010640}. 

Thus, the full marginal log-likelihood is:
\begin{align*}
    \log \mathcal{L}(\mathbf{Y}_n\mid \mathbf{W}_n, \mathbf{X}_n,\theta_y,\theta_w) = \log \mathcal{L}(\mathbf{Y}_n\mid \mathbf{W}_n,\theta_y)+ \log \mathcal{L}(\mathbf{W}_n\mid \mathbf{X}_n, \theta_w).
\end{align*}

An important feature of this framework is that the only hyperparameters requiring estimation are $\theta_y$ and $\theta_w$. This is different from  the DGP in \citet{Sauer02012023}, where the latent layer involves $D$ separate lengthscale parameters ${\theta_{w_1},\dots,\theta_{w_D}}$ due to the assumption of $D$ independent GPs (see Figure~\ref{fig:deepGPmulti}). In our case, the inclusion of the coregionalization matrix $\mathbf{B}_w$ not only enables modeling of dependencies across latent dimensions, but also reduces computational complexity by replacing $D$ separate lengthscale parameters with a single shared $\theta_w$, yielding a more efficient inference procedure via the marginal likelihood.

With the marginal log-likelihood in place, we proceed to posterior inference. The posterior distribution is given by
$$
\pi(\mathbf{W}_n,\theta_w,\theta_y|\mathbf{Y}_n)\propto\mathcal{L}(\mathbf{Y}_n\mid \mathbf{W}_n, \mathbf{X}_n,\theta_y,\theta_w)\pi(\theta_w)\pi(\theta_y).
$$
We first focus on updating the hyperparameters $\theta_w$ and $\theta_y$ using the uniform sliding-window Metropolis–Hastings (MH) scheme \citep{tgp2017, Sauer02012023}. Specifically, the proposed values are drawn as
\begin{align*}
    \theta_w^* \sim \mathrm{Unif}\left(\frac{\ell \theta_w^{(t-1)}}{u}, \frac{u \theta_w^{(t-1)}}{\ell}\right)\quad\text{and}\quad\theta_y^* \sim \mathrm{Unif}\left(\frac{\ell \theta_y^{(t-1)}}{u}, \frac{u \theta_y^{(t-1)}}{\ell}\right),
\end{align*}
with the MH acceptance probabilities:
\begin{equation}\label{eq:MHw}
\alpha_w=\min \left(1, \frac{\mathcal{L}(\mathbf{W}_n\mid \mathbf{X}_n,\theta_w^*)\,\pi(\theta_w^*)}{\mathcal{L}(\mathbf{W}_n\mid \mathbf{X}_n,\theta_w^{(t-1)})\,\pi(\theta_w^{(t-1)})}\times\frac{\theta_w^{(t-1)}}{\theta_w^*}\right),    
\end{equation}
and 
\begin{equation}\label{eq:MHy}
\alpha_y=\min \left(1, \frac{\mathcal{L}(\mathbf{Y}_n\mid \mathbf{W}_n,\theta_y^*)\,\pi(\theta_y^*)}{\mathcal{L}(\mathbf{Y}_n\mid \mathbf{W}_n,\theta_y^{(t-1)})\,\pi(\theta_y^{(t-1)})}\times\frac{\theta_y^{(t-1)}}{\theta_y^*}\right).
\end{equation}
Following the recommendation in \citet{Sauer02012023}, we adopt $\ell = 1$ and $u = 2$ for the proposal bounds.

To sample $\mathbf{W}_n$, we adopt  ESS as in \cite{Sauer02012023}. The procedure begins by drawing a random angle  $\gamma \sim \mathrm{Unif}(0, 2\pi)$  and initializing the bracket bounds as  $\gamma_{\min}=\gamma - 2\pi$ and $\gamma_{\mathrm{max}}=\gamma$. A new proposal is then computed as
\begin{align*}
    \mathbf{W}_n^*=\mathbf{W}_n^{(t-1)}\cos\gamma + \mathbf{W}_n^{\text{prior}}\sin\gamma, 
\end{align*}
where the prior sample $\mathbf{W}^{\mathrm{prior}}_n$ is drawn from the following distribution by leveraging the marginal likelihood \eqref{eq:marglikW}:
$$\mathbf{w}^{\mathrm{prior}}_{nD}\sim\mathcal{N}_{nD}(\mathbf{0}, \hat{\mathbf{B}}_w \otimes K_{\theta_w}(\mathbf{X}_n)),$$
with $\mathbf{w}^{\mathrm{prior}}_{nD}=\text{vec}(\mathbf{W}^{\mathrm{prior}}_n)$. The acceptance probability for the proposed $\mathbf{W}_n^*$ is
\begin{equation}\label{eq:ESSW}
   \alpha_W=\min\left(1,\frac{\mathcal{L}(\mathbf{Y}_n\mid \mathbf{W}_n^*,\theta_y)}{\mathcal{L}(\mathbf{Y}_n\mid \mathbf{W}_n^{(t-1)},\theta_y)}\right). 
\end{equation}
If the proposal is rejected, the bracket is updated by shrinking toward the rejected angle, and a new angle $\gamma$ is sampled from $\mathrm{Unif}(\gamma_{\min}, \gamma_{\max})$. A new proposal is then generated, and this process repeats until a proposal is accepted.

The full Gibbs–ESS–MH sampling procedure for deepICMGP proceeds as follows. After initializing $\theta_y^{(1)}, \theta_w^{(1)}, \mathbf{W}_n^{(1)}$, we iterate for $t = 2, \dots, T$:
\begin{enumerate}
    \item Sample the lengthscale parameters $\theta_w^{(t)}$ and $\theta_y^{(t)}$ using MH acceptance probabilities \eqref{eq:MHw} and \eqref{eq:MHy}, respectively.
    \item Sample the latent layer $\mathbf{W}_n^{(t)}$ using via \eqref{eq:ESSW}.
\end{enumerate}


\subsection{Prediction}\label{sec:prediction}
With the MCMC samples obtained during model training, we are ready to make predictions at new input locations in a multi-output setting. Let $\mathbf{x}_{n+1} \in \mathbb{R}^d$ denote the test input (predictive) location and the corresponding outputs as $\mathbf{y}_{n+1}\in\mathbb{R}^Q$. Suppose that, after training, we have collected posterior samples of the model parameters and latent variables denoted by $\{\theta_y^{(t)}, \theta_w^{(t)}, \mathbf{W}_n^{(t)}\}_{t=1}^T$. The corresponding empirical coregionalization matrices at each iteration are given by $\hat{\mathbf{B}}_y^{(t)}$ and $\hat{\mathbf{B}}_w^{(t)}$.

Let $\mathbf{w}^* \equiv \mathbf{f}_1(\mathbf{x}^*)$ and $\mathbf{y}^* \equiv \mathbf{f}_2(\mathbf{w}^*)$ represent the unknown outputs in the first and second (final) layers, respectively, given the predictive input $\mathbf{x}^*$. Since both $\mathbf{f}_1$ and $\mathbf{f}_2$ are modeled using ICMs, the posterior predictive distributions at each layer follow a multivariate Student-$t$ distribution due to the properties of the conditional multivariate normal distribution \citep{CONTI2010640}. Specifically,
\begin{align}\label{eq:postW}
    \mathbf{w}^{*(t)}\mid\, \mathbf{x}^*, \theta^{(t)}_w, &\mathbf{W}^{(t)}_n, \mathbf{X}_n \sim t_{D}\Big(\boldsymbol{\mu}_w(\mathbf{x}^*), K_{w}(\mathbf{x}^*\mid \mathbf{X}_n)\hat{\mathbf{B}}^{(t)}_w;n \Big),\\
    \boldsymbol{\mu}_w(\mathbf{x}^*) &= \left(\mathbf{I}_D \otimes (K_{\theta^{(t)}_w}(\mathbf{x}^*, \mathbf{X}_n)K_{\theta^{(t)}_w}(\mathbf{X}_n)^{-1})\right)\mathbf{w}^{(t)}_{nD},\nonumber\\
    K_w(\mathbf{x}^*\mid \mathbf{X}_n) &= 1-K_{\theta^{(t)}_w}(\mathbf{x}^*,\mathbf{X}_n)K_{\theta^{(t)}_w}(\mathbf{X}_n)^{-1}K_{\theta^{(t)}_w}(\mathbf{X}_n,\mathbf{x}^*),\nonumber
\end{align} 
and 
\begin{align}
    \mathbf{y}^{*(t)}\mid\, \mathbf{w}^{*(t)}, &\theta^{(t)}_y, \mathbf{W}^{(t)}_n, \mathbf{Y}_n \sim t_{Q}\Big(\boldsymbol{\mu}_y(\mathbf{w}^{*(t)}), K_{y}(\mathbf{w}^{*(t)}\mid \mathbf{W}^{(t)}_n)\hat{\mathbf{B}}^{(t)}_y;n \Big),\label{eq:postY}\\
    \boldsymbol{\mu}_y(\mathbf{w}^{*(t)}) &= \left(\mathbf{I}_Q \otimes (K_{\theta_y}(\mathbf{w}^{*(t)}, \mathbf{W}^{(t)}_n)K_{\theta_y}(\mathbf{W}^{(t)}_n)^{-1})\right)\mathbf{y}_{nQ},\nonumber\\
    K_{y}(\mathbf{w}^{*(t)}\mid \mathbf{W}^{(t)}_n) &=1-K_{\theta^{(t)}_y}(\mathbf{w}^{*(t)},\mathbf{W}^{(t)}_n)K_{\theta^{(t)}_y}(\mathbf{W}^{(t)}_n)^{-1}K_{\theta^{(t)}_y}(\mathbf{W}^{(t)}_n,\mathbf{w}^{*(t)}),\label{eq:postYvar}
\end{align} 
where $\mathbf{I}_D$ and $\mathbf{I}_Q$ denote identity matrices of sizes $D \times D$ and $Q \times Q$, respectively, and $t_{S}(\boldsymbol{\mu},\boldsymbol{\Sigma};n)$ denotes an $S$-dimensional multivariate Student-$t$ distribution with $n$ degrees of freedom, location parameter $\boldsymbol{\mu}$, and scale matrix parameter $\boldsymbol{\Sigma}$. Each kernel matrix $K_{\theta}(\mathbf{x}^*, \mathbf{X}_n)$ is an $1 \times n$ matrix with entries $[K_{\theta}(\mathbf{x}^*, \mathbf{X}_n)]_{1j} = K_{\theta}(\mathbf{x}^*, \mathbf{x}_j)$. While the theoretically correct choice is the Student-$t$ distribution, we find that using a normal distribution yields similar results except when the sample size is very small. Therefore, we use the normal distribution to generate posterior samples in practice.

To generate a posterior sample of the predictive output $\mathbf{y}_{n+1}^{(t)}$, we first sample $\mathbf{w}_{n+1}^{(t)}$  from the conditional distribution in \eqref{eq:postW}, given the predictive input location $\mathbf{x}_{n+1}$. Then, we sample   $\mathbf{y}_{n+1}^{(t)}$ from \eqref{eq:postY}, conditioned on $\mathbf{w}_{n+1}^{(t)}$. The empirical posterior mean and covariance of $\mathbf{y}_{n+1}$ across MCMC samples $\mathcal{T}$ can then be approximated by:
\begin{align*}
    \boldsymbol{\mu}_{y,\mathrm{post}}&=\frac{1}{|\mathcal{T}|}\sum_{t\in \mathcal{T}} \mu_y(\mathbf{w}_{n+1}^{(t)}),\\
    \Sigma_{y, \mathrm{post}} &= \frac{1}{|\mathcal{T}|}\sum_{t\in \mathcal{T}}\left(K_{y}(\mathbf{w}_{n+1}^{(t)}\mid \mathbf{W}^{(t)}_n)\hat{\mathbf{B}}^{(t)}_y +\left(\mu_y(\mathbf{w}_{n+1}^{(t)})-\boldsymbol{\mu}_{y,\mathrm{post}}\right)\left(\mu_y(\mathbf{w}_{n+1}^{(t)})-\boldsymbol{\mu}_{y,\mathrm{post}}\right)^\top\right).
\end{align*}
As discussed in Section~\ref{sec:ICM}, the coregionalization matrices do not influence the posterior means in a conventional ICM model, as seen in the conditional distributions \eqref{eq:postW} and \eqref{eq:postY}; their effect is limited to the posterior covariances. In contrast, our hierarchical deepICMGP model introduces an indirect dependence of the final posterior mean $\boldsymbol{\mu}_{y,\mathrm{post}}$ on the coregionalization matrix $\hat{\mathbf{B}}_w^{(t)}$. This is because $\mathbf{w}_{n+1}^{(t)}$ is sampled from a distribution whose covariance structure is determined by $\hat{\mathbf{B}}_w^{(t)}$, thereby influencing the posterior mean through the two-layer model hierarchy.




\section{Active Learning for deepICMGP}\label{sec:AL}
When data acquisition is expensive, as in training surrogate models for costly computer experiments, it is often infeasible to evaluate the simulator at many input locations. In such cases, it is crucial to select input points that are most informative for learning. Active Learning (AL), or \textit{sequential design}, provides an effective strategy by selecting new inputs that maximize information gain. Unlike one-shot space-filling designs, such as maximin Latin Hypercube Design (LHD) \citep{MORRIS1995381MmLHD}  or MaxPro design \citep{joseph2015maximum}, which spread points uniformly across the input space, AL adapts to the surrogate model’s current uncertainty and focuses sampling where it will be most beneficial. In this section, we develop AL strategies tailored to the proposed deepICMGP model  for multi-output problems. 

Our approach builds on the Active Learning Cohn (ALC) criterion \citep{Cohn1993,seo2000gaussian}, which selects the next design point by maximizing the expected reduction in predictive uncertainty across the input space if that point were added. While ALC is widely used in single-output GP settings, extending it to multi-output models presents new challenges. Specifically, each output dimension may suggest a different next point, leading to inconsistencies in the selection process.

To address this, we propose a unified extension of the ALC criterion that incorporates a D-optimality perspective. Specifically, we select the next input point that maximizes the overall reduction in joint predictive uncertainty by evaluating the change in the \textit{determinant} of the posterior predictive covariance matrix, thus encouraging informative updates across all outputs. Let $\Sigma_n(\mathbf{x})$ denote the posterior predictive covariance matrix of the outputs at location $\mathbf{x}$, based on the current design $\mathbf{X}_n$. After including a candidate point $\mathbf{x}_{n+1}$ into the design, let $\Sigma_{n+1}(\mathbf{x})$ represent the updated predictive covariance at $\mathbf{x}$ based on the augmented design $\mathbf{X}_{n+1} = \mathbf{X}_n \cup {\mathbf{x}_{n+1}}$. The multi-output ALC criterion is then defined as 
\begin{align}\label{eq:alcMulti}
    \Delta \Sigma_n(\mathbf{x}_{n+1}) = \int_{\mathcal{X}} \left(\big|\Sigma_n(\mathbf{x})\big|-\big|\Sigma_{n+1}(\mathbf{x})\big|\right)\, \text{d}\mathbf{x},
\end{align}
where  the determinant quantifies the volume of predictive uncertainty at $\mathbf{x}$, analogous to the D-optimal criterion in experimental design. Here, $\mathcal{X}$ denotes the input space of $\mathbf{x}$. The goal is to select the next input $\mathbf{x}_{n+1}$ that maximizes $\Delta \Sigma_n(\mathbf{x}_{n+1})$, the reduction in the uncertainty (volume) of the predictive covariance matrix, thereby ensuring a globally informative update for all outputs.

As shown in Section~\ref{sec:prediction}, computing the posterior predictive distribution involves warping the input location  through the latent layer value. Accordingly, a candidate input $\mathbf{x}_{n+1}$ is warped to $\mathbf{w}^{(t)}_{n+1}$ under a particular MCMC iteration $t \in \mathcal{T}$. We denote the augmented latent matrix as $\mathbf{W}^{(t)}_{n+1} = \mathbf{W}^{(t)}_n \cup {\mathbf{w}^{(t)}_{n+1}}$, which corresponds to the augmented input locations $\mathbf{X}_{n+1}$. The ALC criterion is then defined based on the posterior covariance of the predicted outputs $\mathbf{y}^{(t)}$ as in \eqref{eq:postY}, yielding
\begin{align*}
    \Delta \Sigma^{(t)}_n(\mathbf{x}_{n+1}) = \int_{\mathcal{W}} \left(\big|K_{y}(\mathbf{w}\mid \mathbf{W}^{(t)}_n)\hat{\mathbf{B}}^{(t)}_y\big|-\big|K_{y}(\mathbf{w}\mid \mathbf{W}^{(t)}_{n+1})\hat{\mathbf{B}}^{(t)}_y\big|\right)\, \text{d}\mathbf{w},
\end{align*}
where $\mathcal{W}$ is the latent  space.

Because $\hat{\mathbf{B}}^{(t)}_y$ is constant with respect to both $\mathbf{w}$ and $\mathbf{w}^{(t)}_{n+1}$ inside the integral, the expression simplifies to
\begin{align*}
\Delta \Sigma^{(t)}_n(\mathbf{x}_{n+1})
    &= \big|\hat{\mathbf{B}}^{(t)}_y\big|\left(\int_{\mathbf{w} \in \mathcal{W}}K_y(\mathbf{w} \mid \mathbf{W}^{(t)}_n)^Q-K_y(\mathbf{w} \mid \mathbf{W}^{(t)}_{n+1})^Q\,d\mathbf{w}\right).
\end{align*}
Since $K_y(\mathbf{w} \mid \mathbf{W}^{(t)}_n)$ does not depend on the new point $\mathbf{x}_{n+1}$ (or equivalently $\mathbf{w}^{(t)}_{n+1}$),  we can define the selection criterion by aggregating over the MCMC samples $t\in\mathcal{T}$ as:
\begin{align*}
    \mathbf{x}^*&=\underset{\mathbf{x}_{n+1} \in \mathcal{X}}{\arg\min} \,\,\text{ALC}(\mathbf{x}_{n+1}),
    \end{align*}
where 
$$
\text{ALC}(\mathbf{x}_{n+1})=\frac{1}{\mathcal{T}}\sum_{t\in\mathcal{T}}\int_{\mathbf{w} \in\mathcal{W}}K_y\left(\mathbf{w} \mid \mathbf{W}^{(t)}_{n} \cup \mathbf{w}^{(t)}_{n+1}\right)^Q\,d\mathbf{w}.
$$

The integral may be tractable for certain kernels over a rectangular domain $\mathcal{W}$, such as the squared exponential kernel in \eqref{eq:sq_exp} \citep{binois2019replication,gramacy2020surrogates}. However, in our case, the latent space $\mathcal{W}$ is generally unknown and not a simple shape. To address this, we follow the approximation strategy in \citet{Sauer02012023}, using a uniform \textit{reference} set in the input space $\mathcal{X}$, denoted $\mathbf{X}_{\text{ref}}$. These reference inputs are mapped to the latent space as $\mathbf{W}^{(t)}_{\text{ref}}$ using the posterior predictive distribution in \eqref{eq:postW}. Thus, using the posterior variance expression from \eqref{eq:postYvar}, the ALC criterion can be approximated by
\begin{align*}
    &\text{ALC}(\mathbf{x}_{n+1})\approx \frac{1}{|\mathcal{T}|} \frac{1}{|\mathbf{X}_{\text{ref}}|}\sum_{t\in\mathcal{T}}\sum_{\mathbf{w} \in \mathbf{W}^{(t)}_{\text{ref}}} \left(1-K_{\theta^{(t)}_y}(\mathbf{w},\mathbf{W}^{(t)}_{n+1})K_{\theta^{(t)}_y}(\mathbf{W}^{(t)}_{n+1})^{-1}K_{\theta^{(t)}_y}(\mathbf{W}^{(t)}_{n+1},\mathbf{w})\right)^Q. 
\end{align*}
In this paper, we use a space-filling design for $\mathbf{X}_{\text{ref}}$ based on a maximin LHD \citep{MORRIS1995381MmLHD}. Note that the selection criterion does not depend on the coregionalization matrix $\hat{\mathbf{B}}^{(t)}_y$, as it is independent of the input $\mathbf{x}$. This implies that the predictive covariance remains fixed across all locations $\mathbf{x}$, which simplifies the computation. In particular, the variance update can be efficiently computed using the partitioned inverse and Sherman--Morrison formula \citep{harville1998matrix}, leading to:
\begin{align*}
    \;\quad K_{\theta^{(t)}_y}(\mathbf{w},\mathbf{W}^{(t)}_{n+1})K_{\theta^{(t)}_y}(\mathbf{W}^{(t)}_{n+1})^{-1}K_{\theta^{(t)}_y}&(\mathbf{W}^{(t)}_{n+1},\mathbf{w})=K_{\theta^{(t)}_y}(\mathbf{w},\mathbf{W}^{(t)}_{n})K_{\theta^{(t)}_y}(\mathbf{W}^{(t)}_{n})^{-1}K_{\theta^{(t)}_y}(\mathbf{W}^{(t)}_{n},\mathbf{w})\\
    &+v\left(K_{\theta^{(t)}_y}(\mathbf{w},\mathbf{W}^{(t)}_{n}) \mathbf{h}\right)^2+2zK_{\theta^{(t)}_y}(\mathbf{w},\mathbf{W}^{(t)}_{n}) \mathbf{h}+v^{-1}z^2,
\end{align*}
where $v=1-K_{\theta^{(t)}_y}(\mathbf{w},\mathbf{W}^{(t)}_{n})K_{\theta^{(t)}_y}(\mathbf{W}^{(t)}_{n})^{-1}K_{\theta^{(t)}_y}(\mathbf{W}^{(t)}_{n},\mathbf{w})$, $\mathbf{h}=-v^{-1}K_{\theta^{(t)}_y}(\mathbf{W}^{(t)}_{n})^{-1}K_{\theta^{(t)}_y}(\mathbf{W}^{(t)}_{n},\mathbf{w})$, and $z=K_{\theta^{(t)}_y}(\mathbf{w}_{n+1}, \mathbf{w})$. This yields a variance update with computational cost $\mathcal{O}(n^2)$. More details can be found in the supplementary materials of \cite{Sauer02012023} and Chapter 6 of \cite{gramacy2020surrogates}.

\section{Numerical Study}\label{sec:numerical}
This section presents a series of experiments to evaluate the performance of the proposed deepICMGP model. In Section~\ref{sec:num_result}, we assess predictive performance across various synthetic examples, and examine the effectiveness of active learning strategies in Section~\ref{sec:num_alc}.

\subsection{Predictive Performance on Synthetic Examples}\label{sec:num_result}

We begin by introducing a deep extension of the Convolved Gaussian Process model in Section~\ref{sec:convdgp}, which will serve as a competitor in this numerical study. 

\subsubsection{Deep Convolved Multi-Output Gaussian Process}\label{sec:convdgp}

In addition to ICM, another widely used MOGP framework is the Convolved Gaussian Process (CvGP) \citep{alvarez2008, JMLR:v12:alvarez11a}. In CvGPs, each output is expressed as the convolution of a smoothing kernel with a latent function drawn from a GP. This formulation provides a flexible way to model dependencies across outputs and has been applied to a variety of problems. For example, \citet{negative2022} used CvGPs to mitigate negative transfer effects, and \citet{lin2022multi} applied them to multi-fidelity modeling.

Specifically, each output function \( f_q(\mathbf{x}) \) is modeled using $R$ shared latent functions as follows:
\begin{align*}
f_q(\mathbf{x}) = \sum_{r=1}^R \int_{\mathbb{R}^d} k^r_q(\mathbf{x} - \mathbf{z})\, u^r(\mathbf{z}) \, d\mathbf{z},
\end{align*}
where \( k^r_q \) is a smoothing kernel and each latent function \( u^r\) is drawn independently from a GP. If both the smoothing kernel and the kernel of the latent GP  \( u^r\) are squared exponential kernels, the integral can be computed in closed form, yielding a GP with a simplified kernel function. Further details can be found in \citet{JMLR:v12:alvarez11a}.


To ensure a fair comparison, we develop a deep extension of this framework, referred to as Deep Convolved Gaussian Process (deepCvGP), which mirrors the architecture of deepICMGP by connecting each layer through a CvGP.

\subsubsection{Competitor Models}

We compare \texttt{deepICMGP} against a diverse set of MOGP approaches. Specifically, we consider the following models:

\begin{itemize}
    \item Deep Convolved Gaussian Process (\texttt{deepCvGP}) introduced in Section \ref{sec:convdgp}.
    \item Independent Single-Output Deep Gaussian Process (\texttt{DGP.Ind}): Assumes each output is independent and modeled by a deep GP as in \cite{Sauer02012023}, implemented via the \textsf{R} package \textsf{deepgp} \citep{booth2024deepgp}.
\item {Multi-Output Deep Gaussian Process} (\texttt{DGP.Multi}): A deep GP where the last layer connects multiple independent GPs, implemented via \textsf{dgpsi} \citep{Ming03042023} in \textsf{Python}.

\item {Independent Single-Output Gaussian Process} (\texttt{GP.Ind}): Assumes each output is independent and modeled by a standard GP, implemented via the \textsf{R} package \textsf{laGP} \citep{gramacy2016lagp}.

\item {Linear Model of Coregionalization} (\texttt{LMC}): Proposed by \cite{NIPS2007_bonilla} and implemented via \textsf{GPy} \citep{gpy2014} in \textsf{Python}.
\item {Convolved Gaussian Process} (\texttt{CvGP}): Proposed by \cite{alvarez2008,JMLR:v12:alvarez11a} and implemented via \textsf{mogptk} \citep{DEWOLFF202149} in \textsf{Python}.

\item {Cross Spectral Mixture} (\texttt{CSM}): Proposed by \cite{CSM} and implemented via \textsf{mogptk} \citep{DEWOLFF202149} in \textsf{Python}.
\item {Multi-Output Spectral Mixture} (\texttt{MOSM}): Proposed by \cite{MOSM} and implemented via \textsf{mogptk} \citep{DEWOLFF202149} in \textsf{Python}.
\item {Spectral Mixture Linear Model of Coregionalization} (\texttt{SMLMC}): Proposed by \cite{goovaerts1997geostatistics} and \cite{spectralkerenl2013} and implemented via \textsf{mogptk} \citep{DEWOLFF202149} in \textsf{Python}.
\end{itemize}

\subsubsection{Simulation Setup}\label{sec:numsetup}
We consider eight synthetic examples that are commonly used as benchmarks in the literature to evaluate prediction performance. These synthetic functions are provided in the Appendix \ref{sup: functions}. The input dimension and the training/testing sample sizes for each example are summarized in Table~\ref{tab:example_info}. Both training and testing datasets were generated using the maximin LHD \citep{MORRIS1995381MmLHD}. 

\begin{table}[]
    \centering
    \begin{tabularx}{\textwidth}{c Y Y Y Y Y Y Y Y Y Y}
    \toprule

            &  &   &  Damped &   & &  &  &  \\
        & Forrester & Convolved  &  Wave & Pedikaris  &Branin & MOP2 & Currin & Park \\
    \midrule
        $d$ & 1 & 1 & 1 & 1& 2 & 2 & 2 & 4 \\
        $Q$ & 2 & 3 & 3 & 2& 3 & 2 & 2 & 2 \\
        $n_{\text{train}}$  & 9 & 10 & 15 & 12 & 30 & 30 & 30 & 60 \\
        $n_{\text{test}}$   & 100 & 100 & 100 & 100& 500 & 500 & 500 & 1000 \\
    \bottomrule
    \end{tabularx}
    \caption{Input dimensions and sample sizes for the synthetic examples.}
    \label{tab:example_info}
\end{table}

To ensure a fair comparison across models, we applied the same sampling strategy and inference configuration wherever applicable. For \texttt{deepICMGP}, \texttt{deepCvGP}, and \texttt{DGP.Ind}, the MCMC procedure was run for 5000 iterations, with a burn-in of 1000 and a thinning interval of 2, resulting in 2000 posterior samples used for prediction. Each of these models was implemented with two layers, and the number of GP nodes in each layer was set to $\max(d, Q)$. An exception is \texttt{deepCvGP}, for which we fixed the number of nodes to $Q$ based on preliminary experiments indicating improved numerical stability under this setting. For the \texttt{DGP.Multi} model, following the recommendation of \cite{Ming03042023}, the number of nodes in each layer was set to $d$.

Other models followed similar procedures adapted to their respective architectures. For the LMC model, the \texttt{L-BFGS-B} optimizer \citep{L-BFGS-B} was used to update hyperparameters during training. For CvGP, CSM, MOSM, and SMLMC, model training was performed over 5000 iterations using the \texttt{Adam} optimizer \citep{Adam2014}, with hyperparameters initialized via the \texttt{Lomb-Scargle} periodogram \citep{Lomb1976,Scargle1982}. 

All experiments were conducted on a high-performance computing cluster (HPCC), using one core per task with 32~GB of RAM allocated per core.


\subsubsection{Prediction Performance Metrics}
We assess the predictive performance of each method using three criteria:

\begin{enumerate}
    \item Root Mean Square Error (RMSE): Evaluates the accuracy of the predictive mean for each individual output.
    \item Continuous Ranked Probability Score (CRPS) \citep{Gneiting01032007}: Evaluates both the predictive mean and variance for each output.
    \item Multivariate Log Score (Score) \citep{Dawid1999Mah, Gneiting01032007}: A proper scoring rule that incorporates the full predictive covariance structure across outputs.
\end{enumerate}

Let $\boldsymbol{\mu}_i$ and $\boldsymbol{\Sigma}_i$ denote the predictive mean vector and covariance matrix for the $i$-th test input, respectively. The diagonals of $\boldsymbol{\Sigma}_i$ correspond to the variances for each output, denoted $\sigma^2_{i,1}, \ldots, \sigma^2_{i,Q}$. Let $\mathbf{y}_i$ be the true observed output for the $i$-th test input. The performance metrics are defined as follows:
\begin{align*}
&\text{RMSE}_q =  \sqrt{ \frac{1}{n_{\text{test}}}\sum_{i=1}^{n_{\text{test}}} (y_{i,q} - \mu_{i,q})^2 },\quad q=1,\ldots,Q, \\
&\text{CRPS}_q = \frac{1}{n_{\text{test}}} \sum_{i=1}^{n_{\text{test}}} \sigma_{i,q}^2 \left[ \frac{1}{\pi} - 2\phi\left( \frac{y_{i,q} - \mu_{i,q}}{\sigma_{i,q}} \right) - \frac{y_{i,q} - \mu_{i,q}}{\sigma_{i,q}} \left( 2\Phi\left( \frac{y_{i,q} - \mu_{i,q}}{\sigma_{i,q}}\right) - 1 \right) \right],\\
&\text{Multivariate Score}=\text{median}_{i=1,\ldots,n_{\text{test}}}  \{-\log |\boldsymbol{\Sigma}_i|-(\mathbf{y}_i-\boldsymbol{\mu})^\top\boldsymbol{\Sigma}_i^{-1}(\mathbf{y}_i-\boldsymbol{\mu})\},
\end{align*}
$\phi$ and $\Phi$ denote the standard normal probability density and cumulative distribution functions, respectively.

Lower values of RMSE and CRPS indicate better predictive accuracy and uncertainty quantification for individual outputs. For the multivariate Score, a higher value indicates better predictive performance across outputs. Note that we found that the distribution of log scores across test points is often highly skewed and sensitive to outliers. To ensure a more robust comparison, we report the median of the Score values instead of  the mean. 

Note that methods implemented in the \textsf{Python} library \textsf{mogptk}, including \texttt{CvGP}, \texttt{CSM}, \texttt{MOSM}, and \texttt{SMLMC}, do not report Score values, as they do not return the predictive covariance matrix, to the best of our knowledge.

\subsubsection{Prediction Results}
\begin{figure}[htbp]
  \centering
  \begin{subfigure}[t]{0.49\linewidth}
    \centering
    \includegraphics[width=0.95\linewidth]{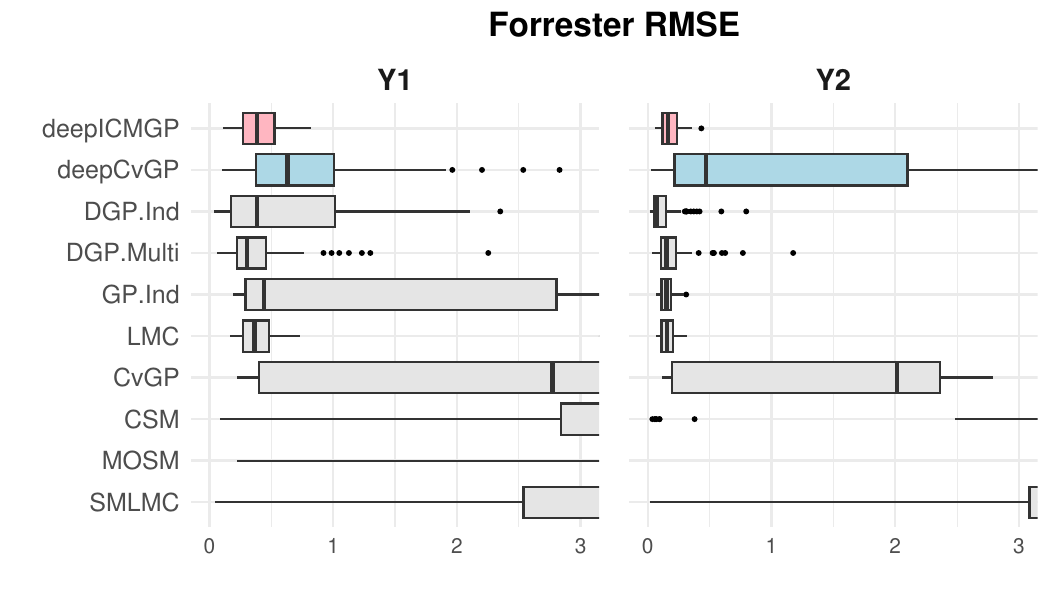}
  \end{subfigure}
  \hfill
  \begin{subfigure}[t]{0.49\linewidth}
    \centering
    \includegraphics[width=0.95\linewidth]{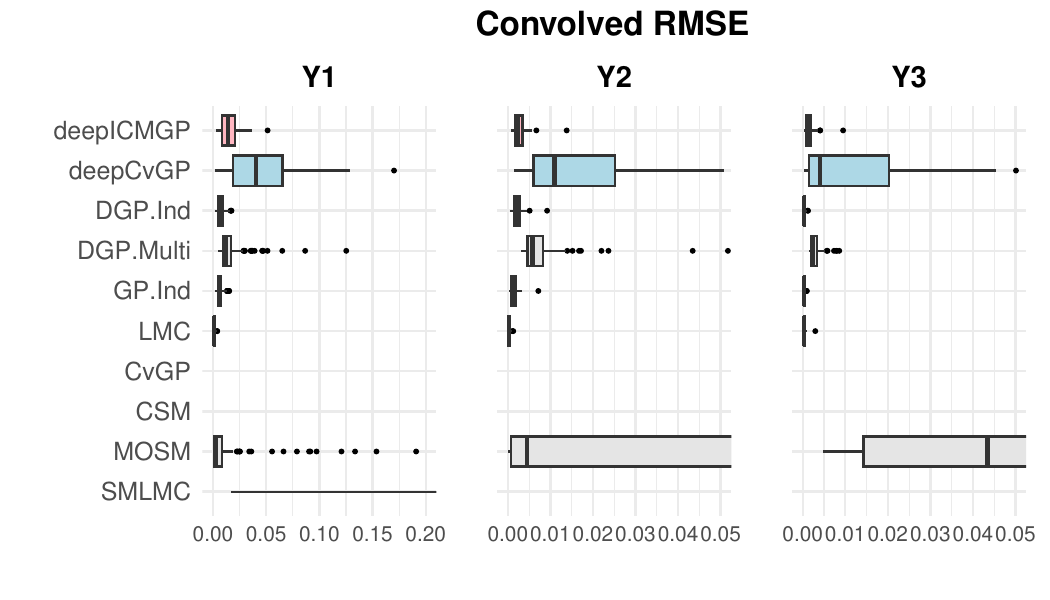}
  \end{subfigure}
  \vskip 0.1\baselineskip
  \begin{subfigure}[t]{0.49\linewidth}
    \centering
    \includegraphics[width=0.95\linewidth]{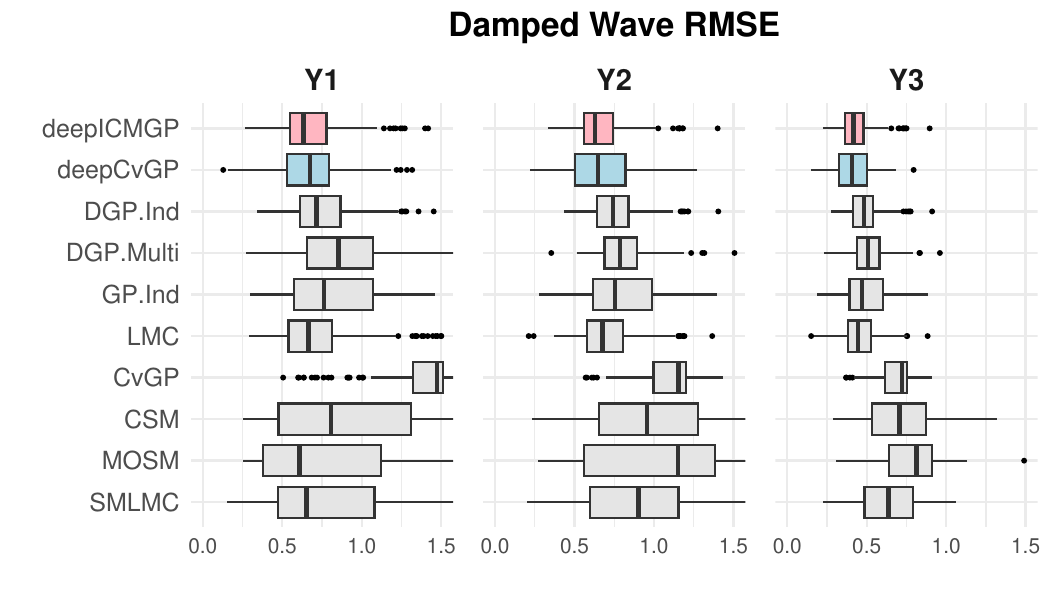}
  \end{subfigure}
  \hfill
  \begin{subfigure}[t]{0.49\linewidth}
    \centering
    \includegraphics[width=0.95\linewidth]{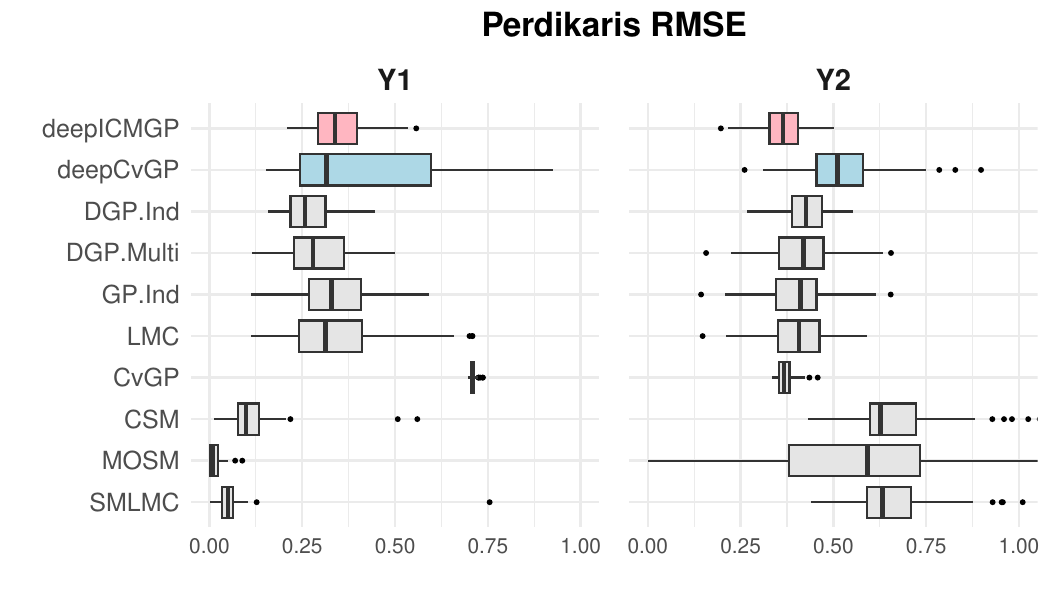}
  \end{subfigure}
  \vskip 0.1\baselineskip
  \begin{subfigure}[t]{0.49\linewidth}
    \centering
    \includegraphics[width=0.95\linewidth]{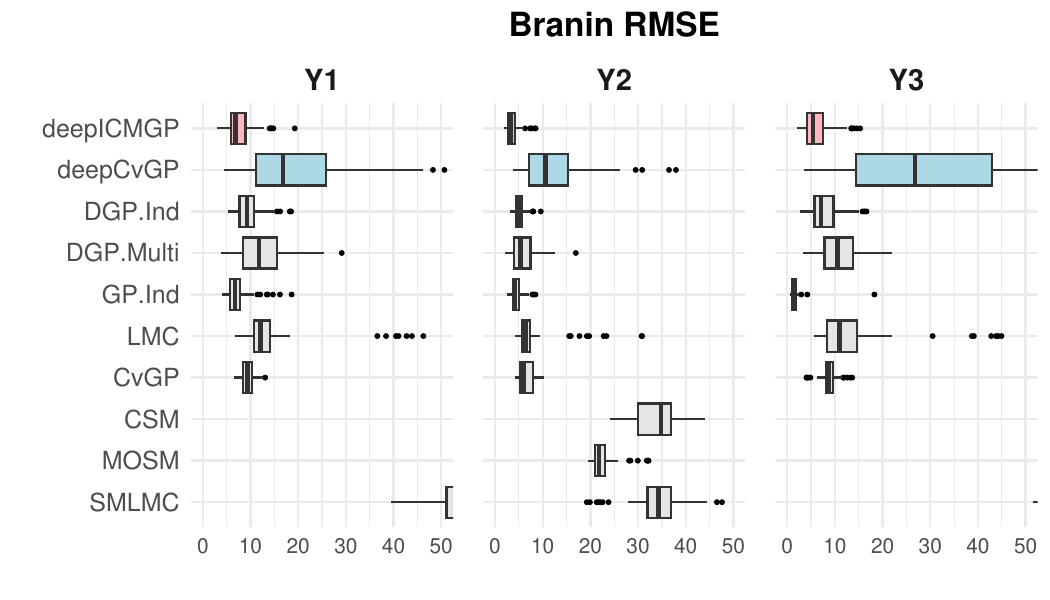}
  \end{subfigure}
  \hfill
  \begin{subfigure}[t]{0.49\linewidth}
    \centering
    \includegraphics[width=0.95\linewidth]{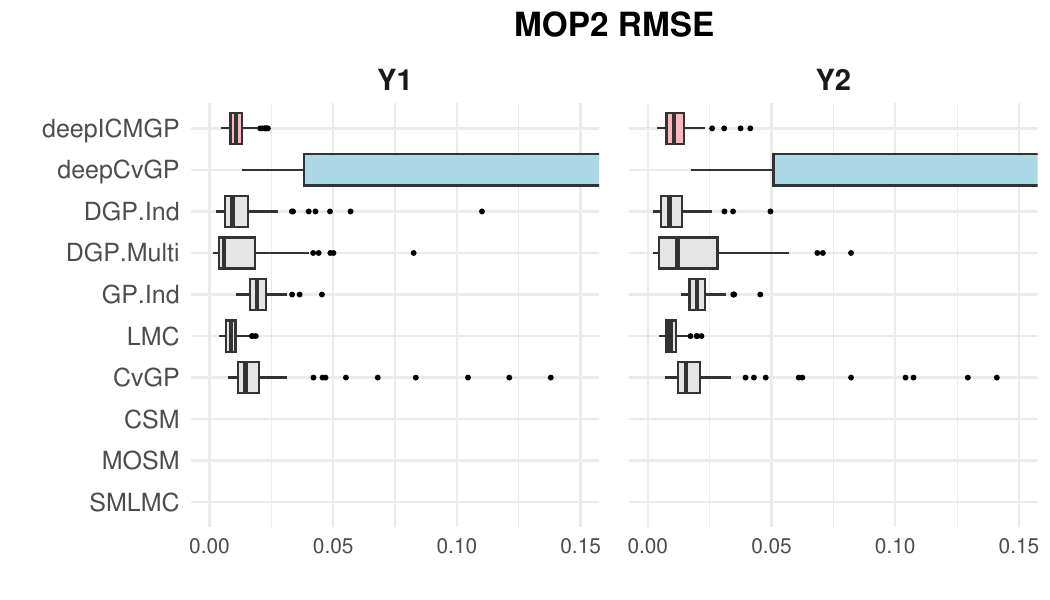}
  \end{subfigure}
  \vskip 0.1\baselineskip
  \begin{subfigure}[t]{0.49\linewidth}
    \centering
    \includegraphics[width=0.95\linewidth]{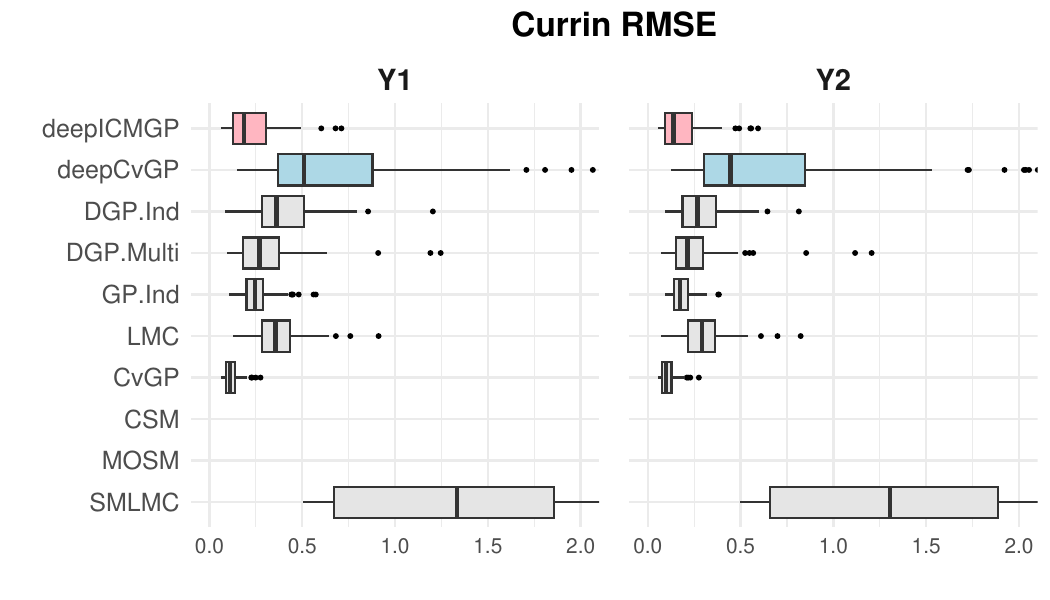}
  \end{subfigure}
  \hfill
  \begin{subfigure}[t]{0.49\linewidth}
    \centering
    \includegraphics[width=0.95\linewidth]{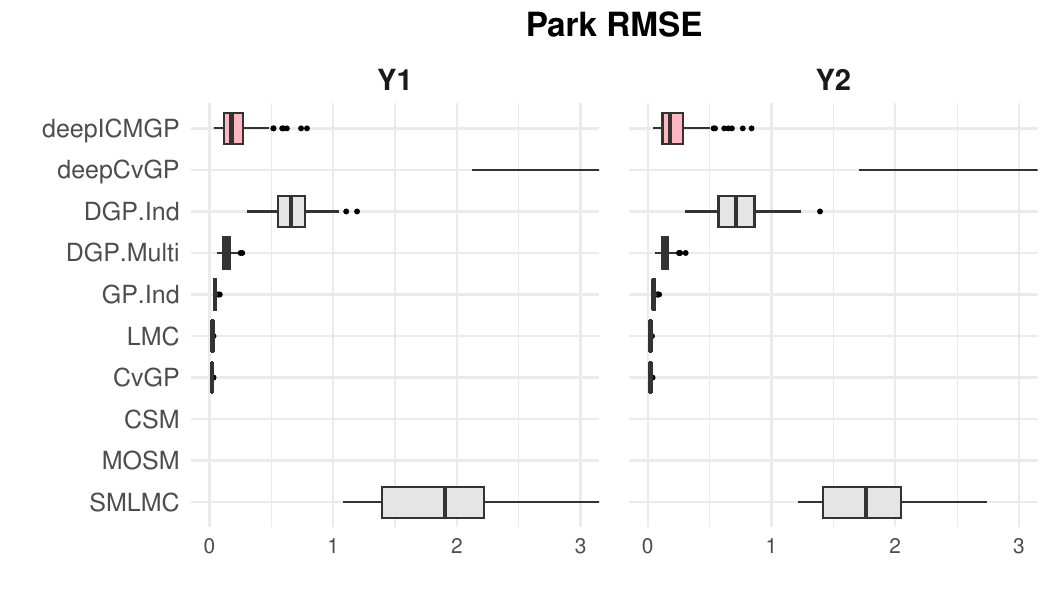}
  \end{subfigure}
  
  \caption{RMSE comparison of \texttt{deepICMGP} and competing methods across all examples. Boxplots summarize results from 100 repetitions with different training samples. For clarity, some extreme boxplots are omitted as their values exceed the displayed range.}
  \label{fig:RMSE_res}
\end{figure}

\begin{figure}[htbp]
  \centering
  \begin{subfigure}[t]{0.49\linewidth}
    \centering
    \includegraphics[width=0.95\linewidth]{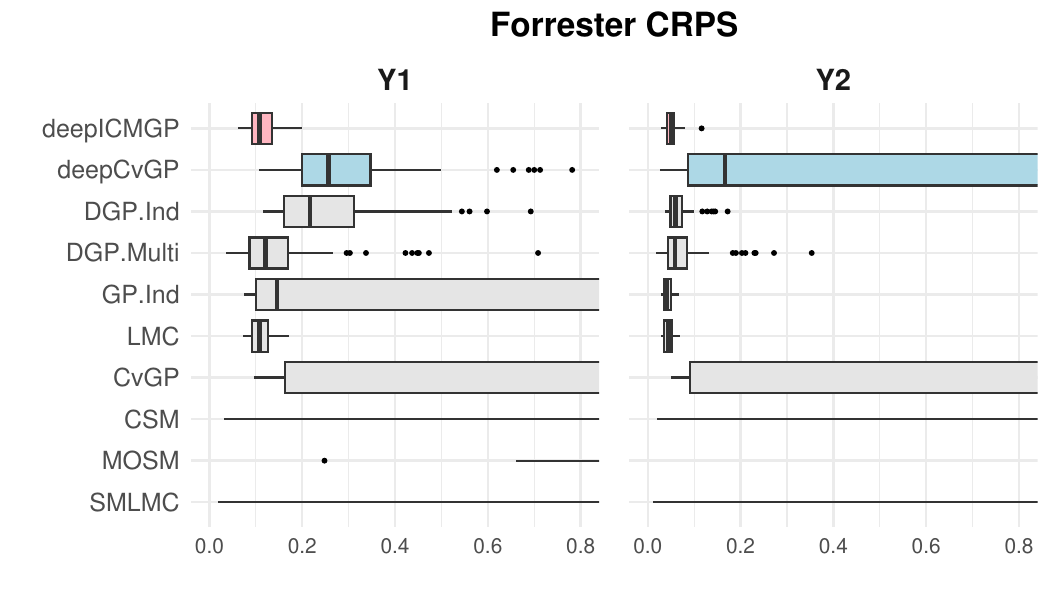}
  \end{subfigure}
  \hfill
  \begin{subfigure}[t]{0.49\linewidth}
    \centering
    \includegraphics[width=0.95\linewidth]{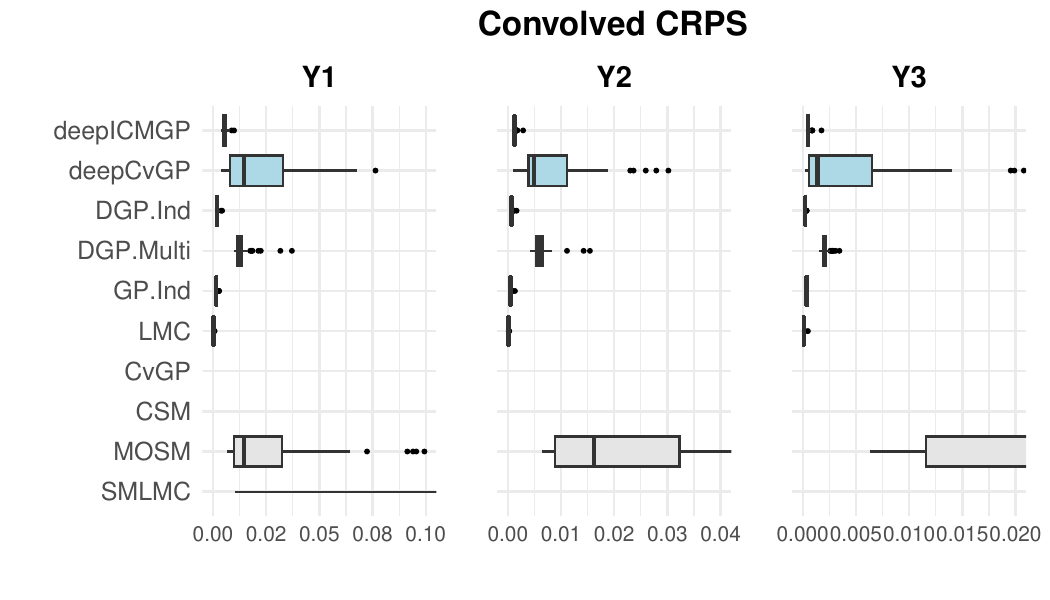}
  \end{subfigure}
  \vskip 0.1\baselineskip
  \begin{subfigure}[t]{0.49\linewidth}
    \centering
    \includegraphics[width=0.95\linewidth]{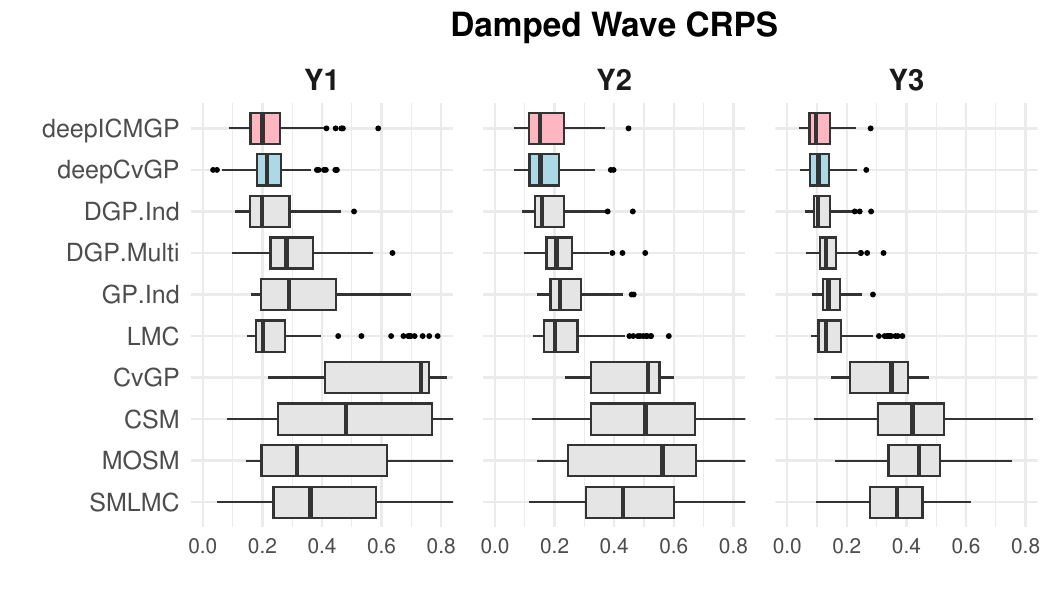}
  \end{subfigure}
  \hfill
  \begin{subfigure}[t]{0.49\linewidth}
    \centering
    \includegraphics[width=0.95\linewidth]{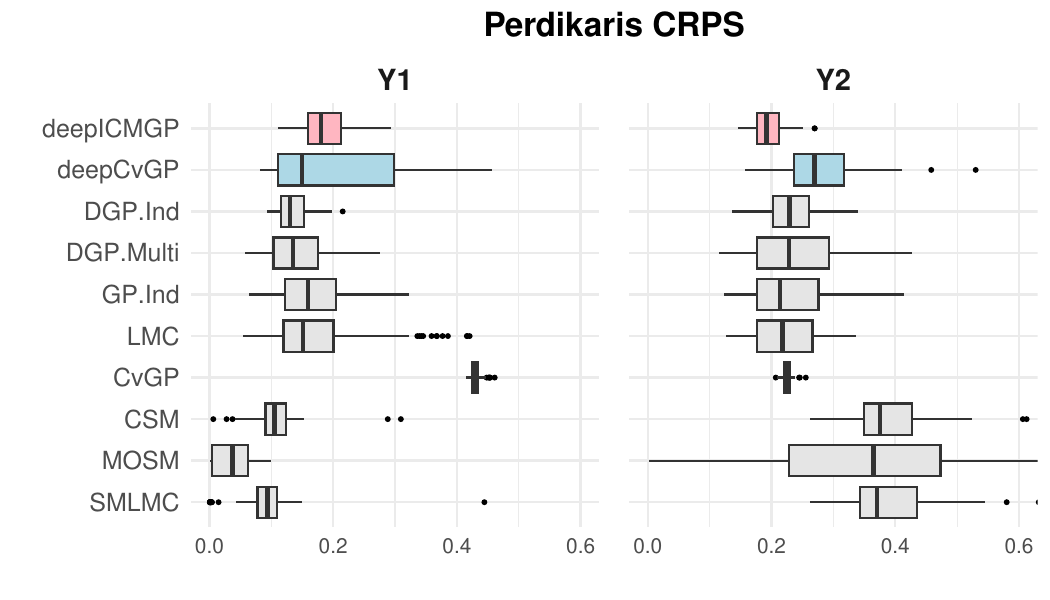}
  \end{subfigure}
  \vskip 0.1\baselineskip
  \begin{subfigure}[t]{0.49\linewidth}
    \centering
    \includegraphics[width=0.95\linewidth]{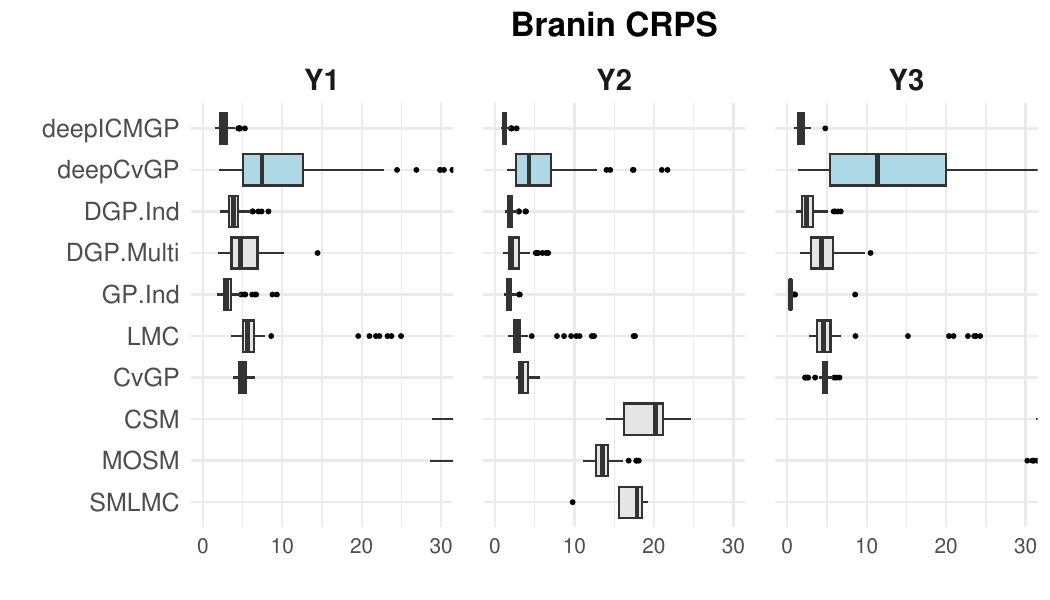}
  \end{subfigure}
  \hfill
  \begin{subfigure}[t]{0.49\linewidth}
    \centering
    \includegraphics[width=0.95\linewidth]{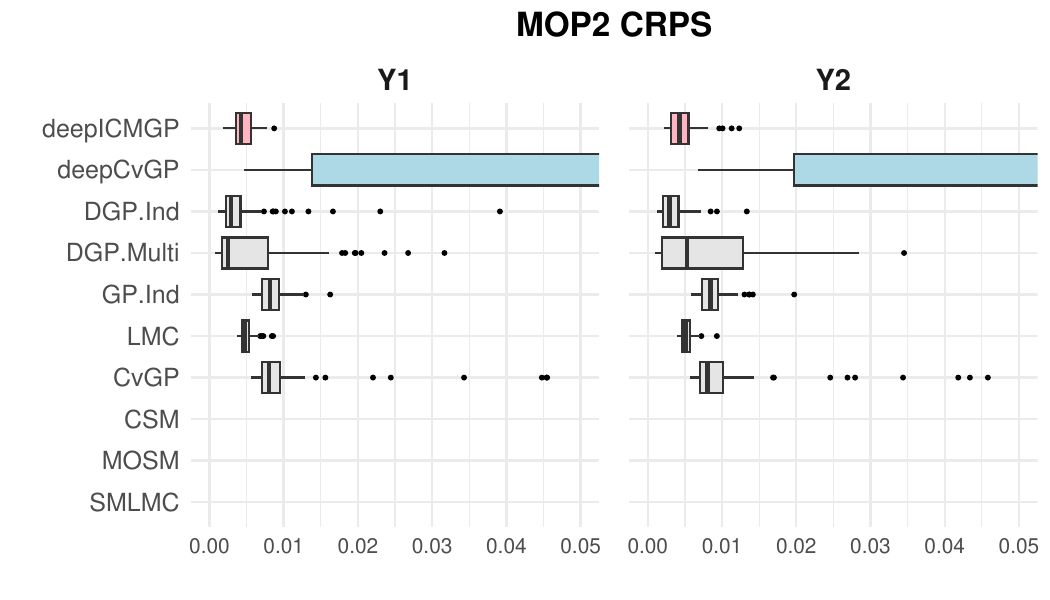}
  \end{subfigure}
  \vskip 0.1\baselineskip
  \begin{subfigure}[t]{0.49\linewidth}
    \centering
    \includegraphics[width=0.95\linewidth]{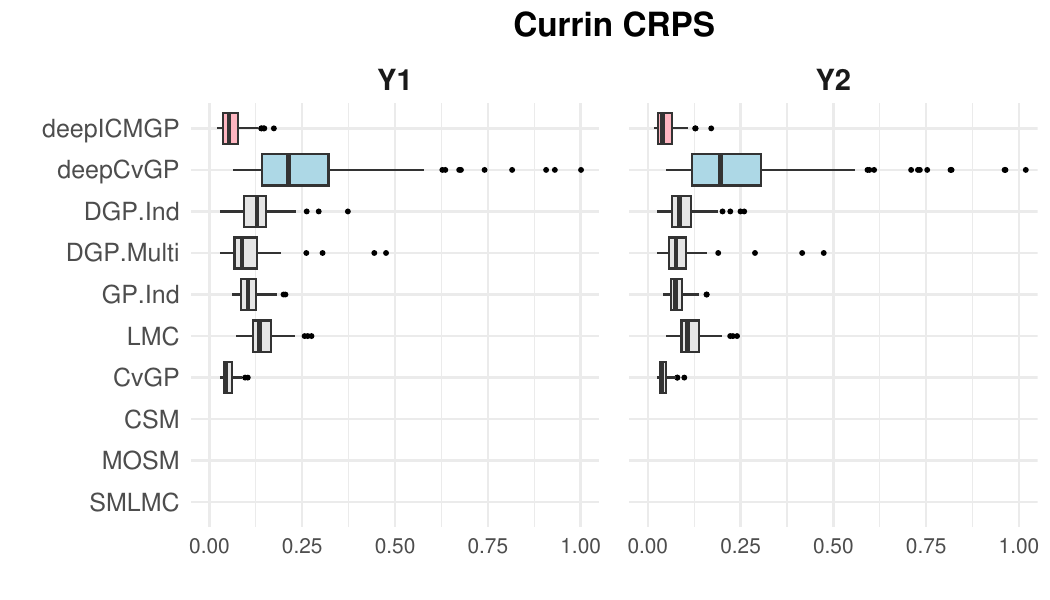}
  \end{subfigure}
  \hfill
  \begin{subfigure}[t]{0.49\linewidth}
    \centering
    \includegraphics[width=0.95\linewidth]{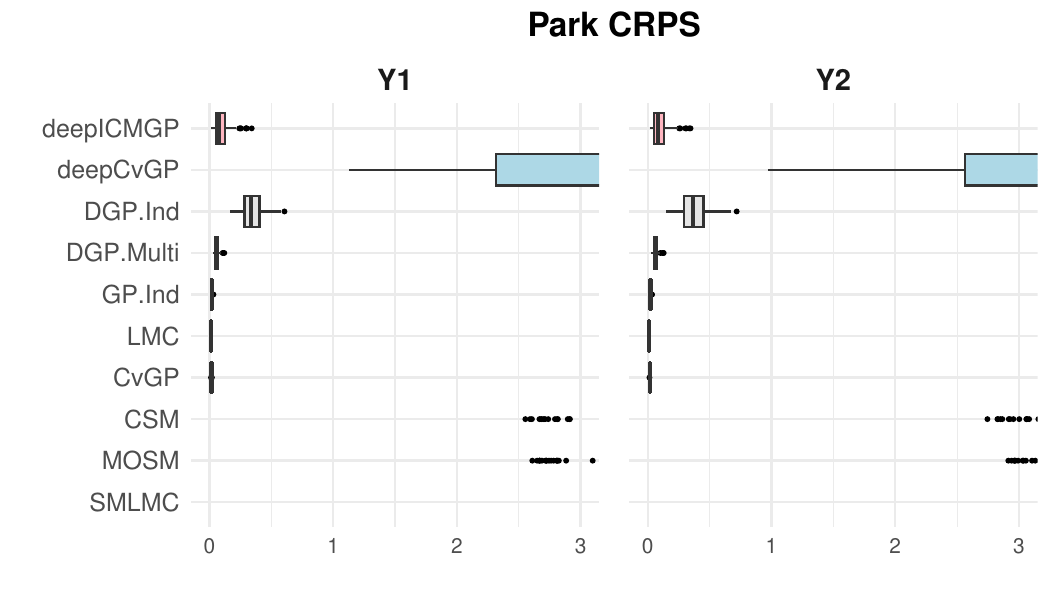}
  \end{subfigure}
  
  \caption{CRPS comparison of \texttt{deepICMGP} and competing methods across all examples. Boxplots summarize results from 100 repetitions with different training samples. For clarity, some extreme boxplots are omitted as their values exceed the displayed range.}
  \label{fig:CRPS_res}
\end{figure}

\begin{figure}[htbp]
  \centering
  \begin{subfigure}[t]{0.49\linewidth}
    \centering
    \includegraphics[width=0.95\linewidth]{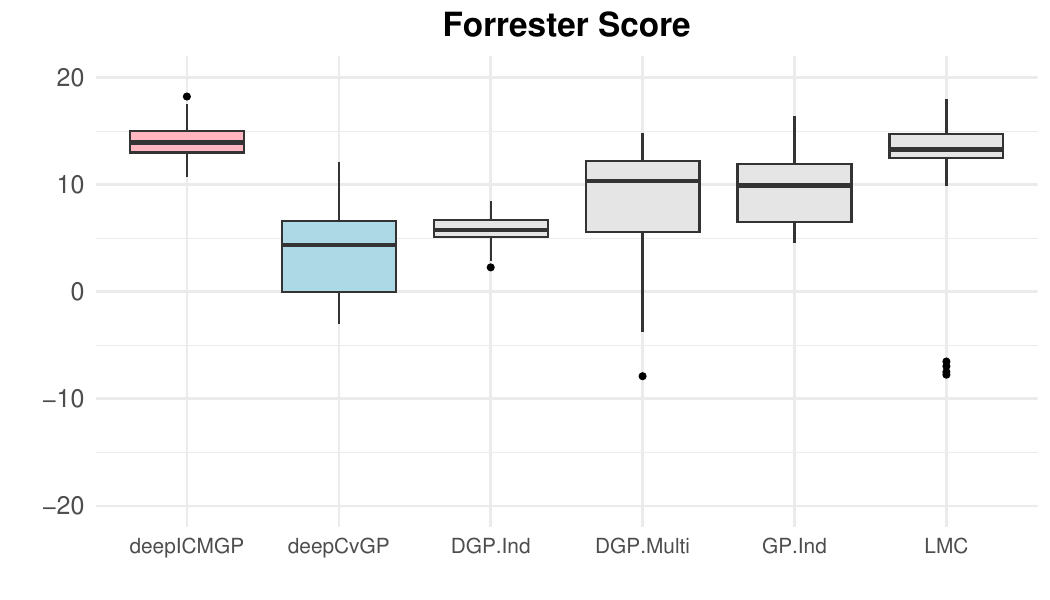}
  \end{subfigure}
  \hfill
  \begin{subfigure}[t]{0.49\linewidth}
    \centering
    \includegraphics[width=0.95\linewidth]{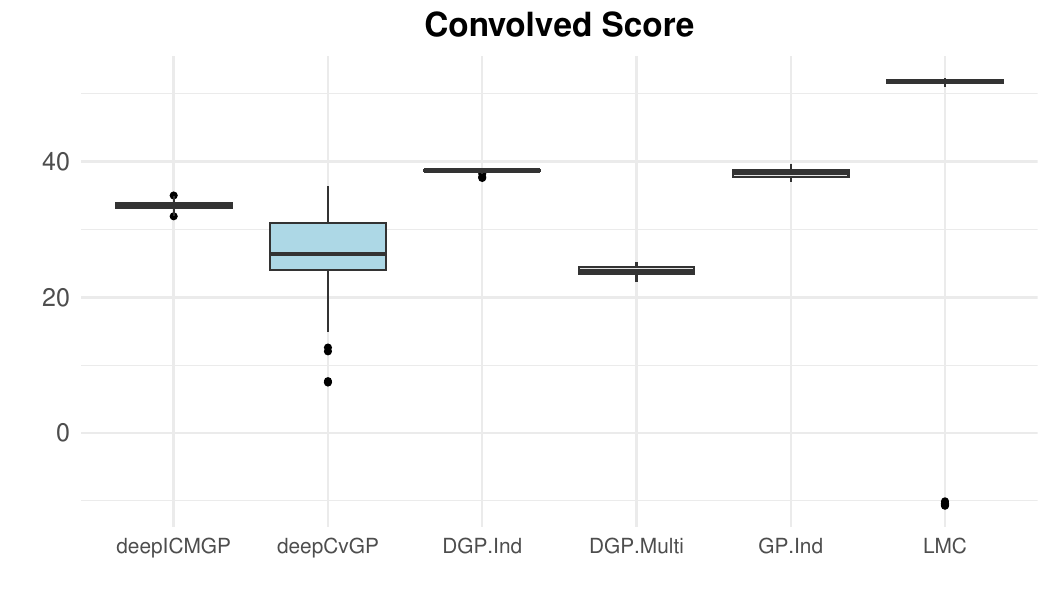}
  \end{subfigure}
  \vskip 0.1\baselineskip
  \begin{subfigure}[t]{0.49\linewidth}
    \centering
    \includegraphics[width=0.95\linewidth]{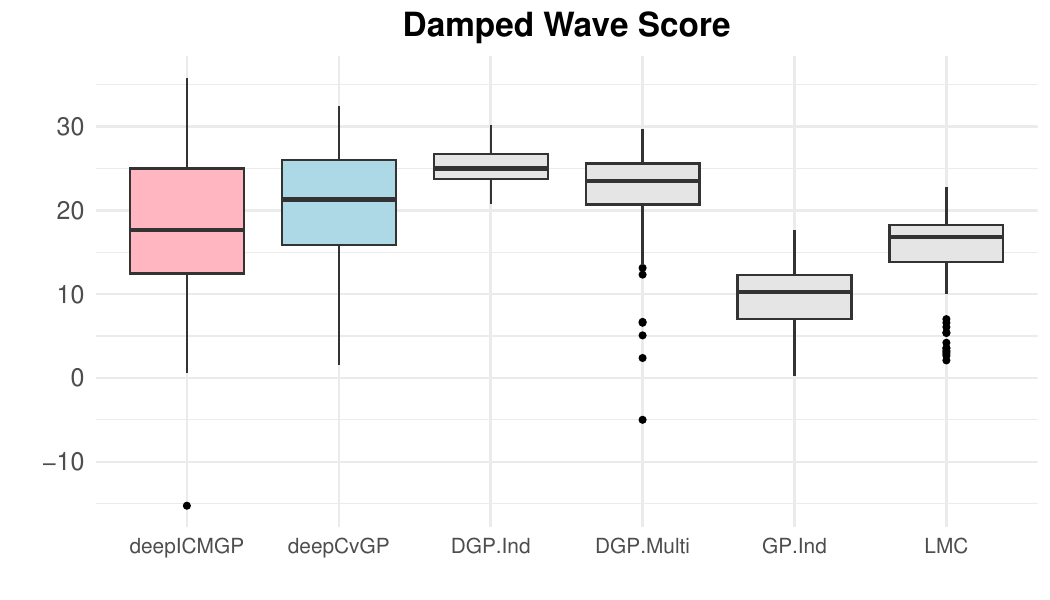}
  \end{subfigure}
  \hfill
  \begin{subfigure}[t]{0.49\linewidth}
    \centering
    \includegraphics[width=0.95\linewidth]{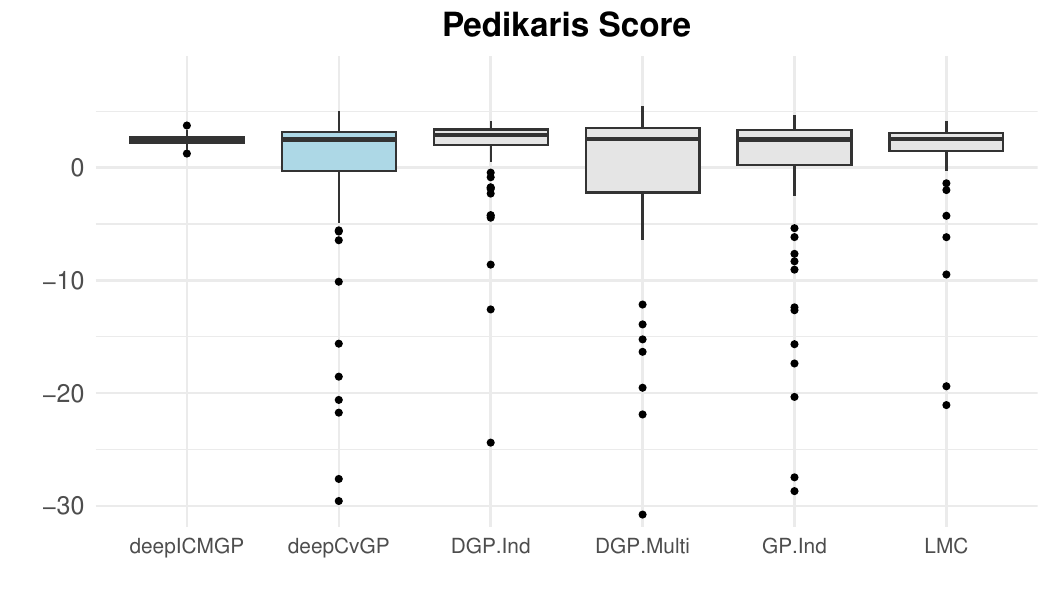}
  \end{subfigure}
  \vskip 0.1\baselineskip
  \begin{subfigure}[t]{0.49\linewidth}
    \centering
    \includegraphics[width=0.95\linewidth]{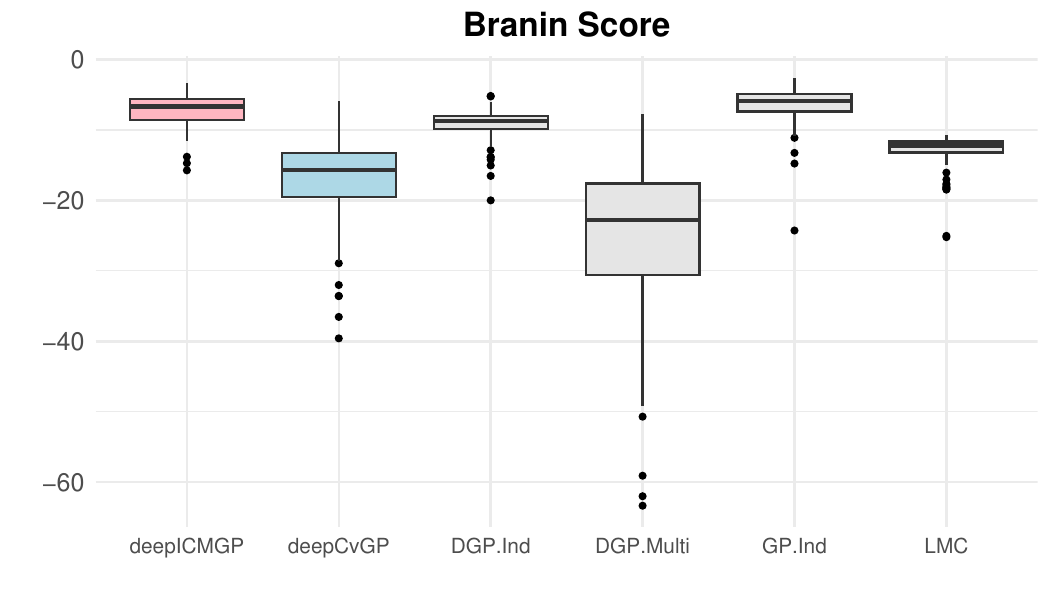}
  \end{subfigure}
  \hfill
  \begin{subfigure}[t]{0.49\linewidth}
    \centering
    \includegraphics[width=0.95\linewidth]{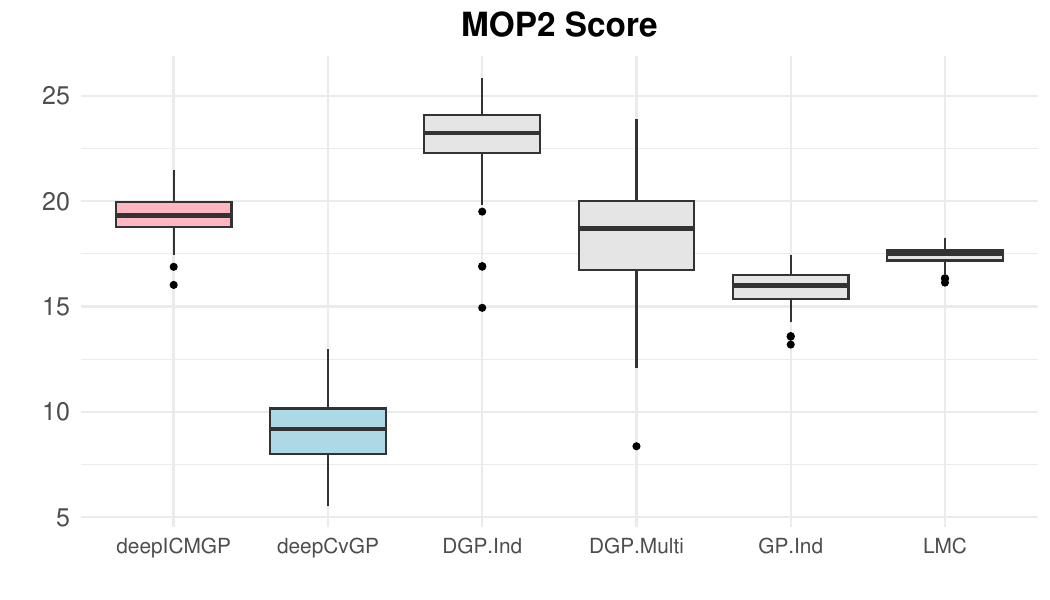}
  \end{subfigure}
  \vskip 0.1\baselineskip
  \begin{subfigure}[t]{0.49\linewidth}
    \centering
    \includegraphics[width=0.95\linewidth]{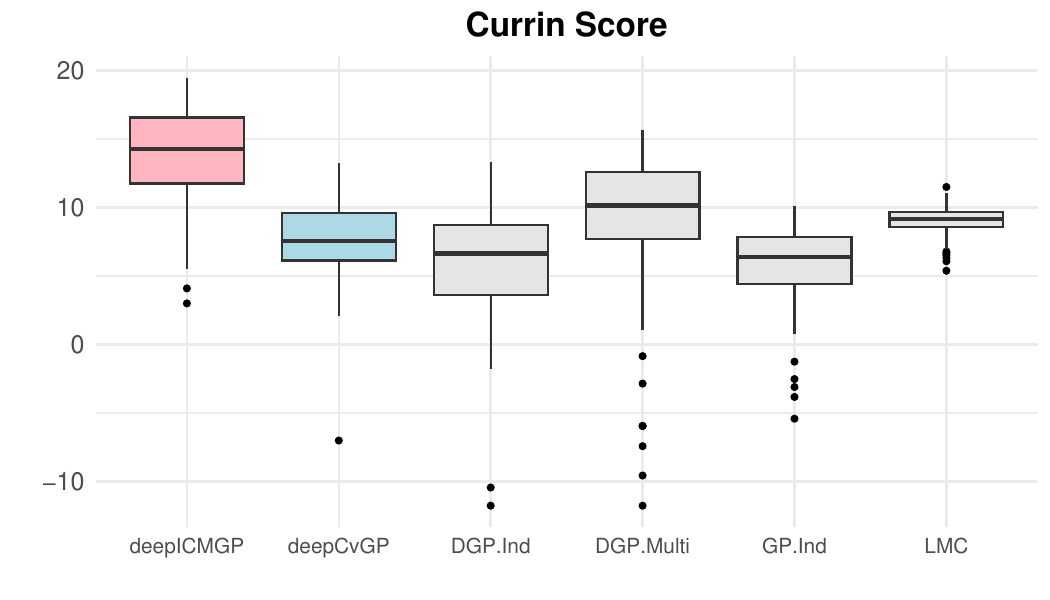}
  \end{subfigure}
  \hfill
  \begin{subfigure}[t]{0.49\linewidth}
    \centering
    \includegraphics[width=0.95\linewidth]{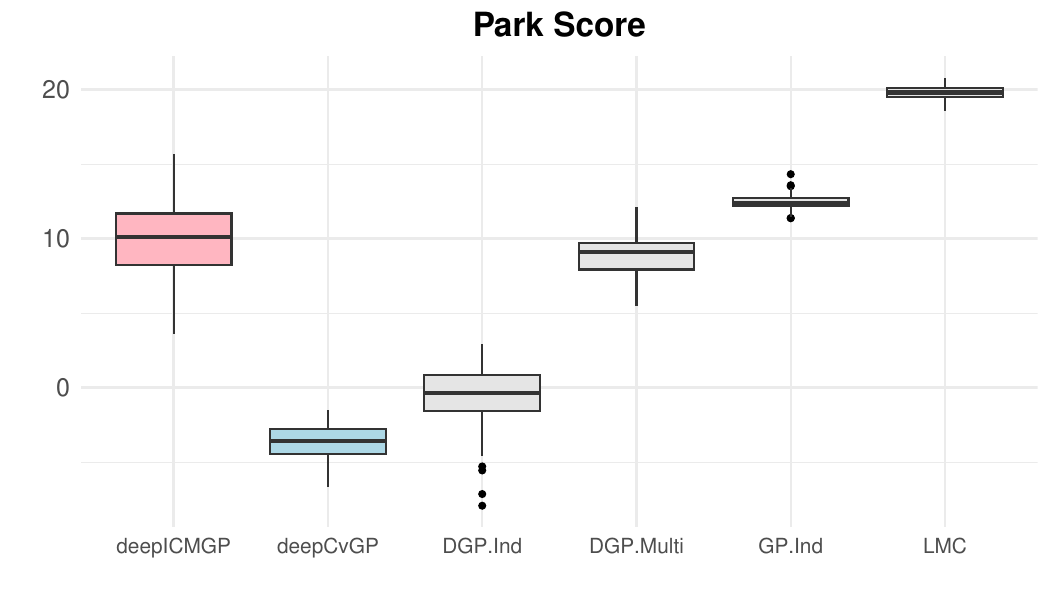}
  \end{subfigure}
  
  \caption{Multivariate Score comparison of \texttt{deepICMGP} and competing methods across all examples. Boxplots summarize results from 100 repetitions with different training samples. }
  \label{fig:score_res}
\end{figure}

\begin{figure}[htbp]
  \centering
  \begin{subfigure}[t]{0.49\linewidth}
    \centering
    \includegraphics[width=0.95\linewidth]{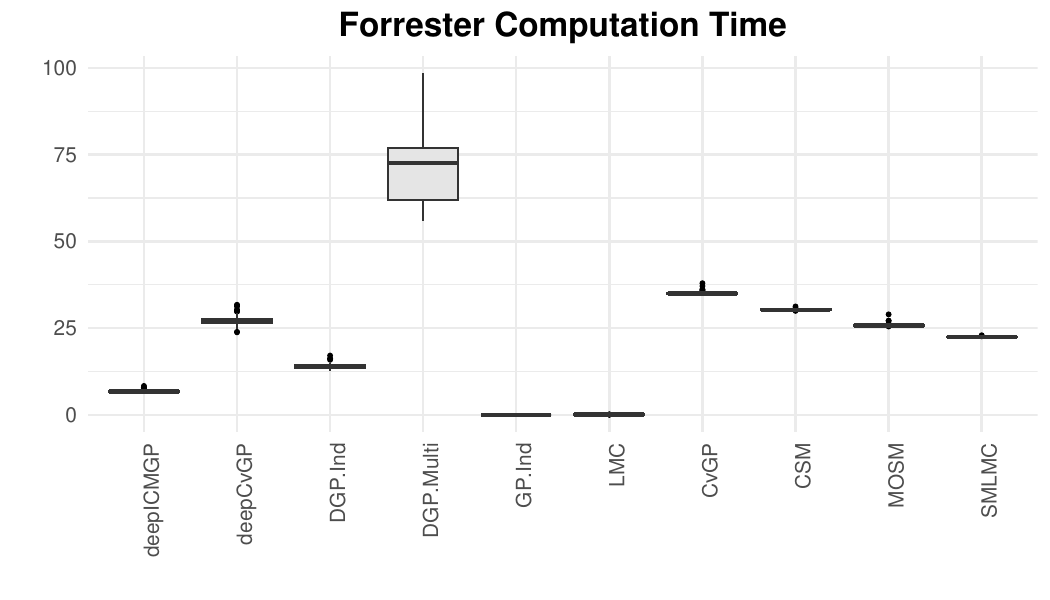}
  \end{subfigure}
  \hfill
  \begin{subfigure}[t]{0.49\linewidth}
    \centering
    \includegraphics[width=0.95\linewidth]{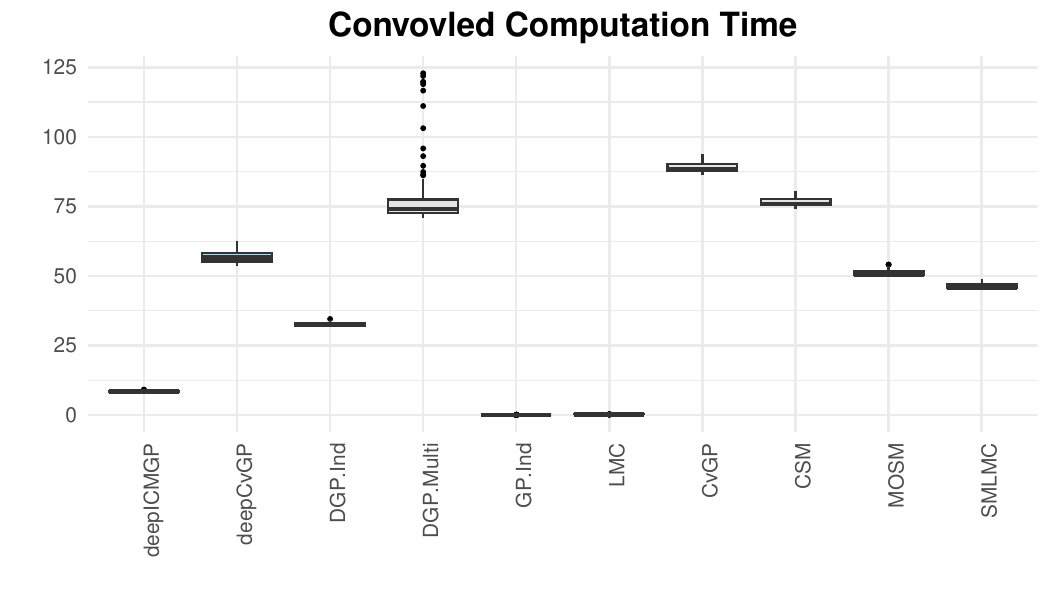}
  \end{subfigure}
  \vskip 0.1\baselineskip
  \begin{subfigure}[t]{0.49\linewidth}
    \centering
    \includegraphics[width=0.95\linewidth]{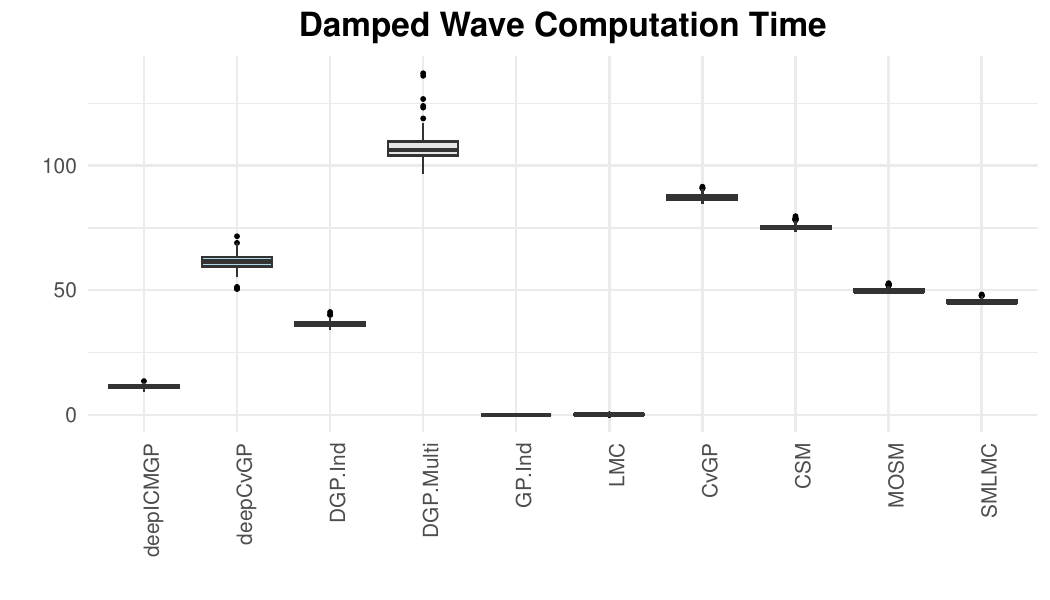}
  \end{subfigure}
  \hfill
  \begin{subfigure}[t]{0.49\linewidth}
    \centering
    \includegraphics[width=0.95\linewidth]{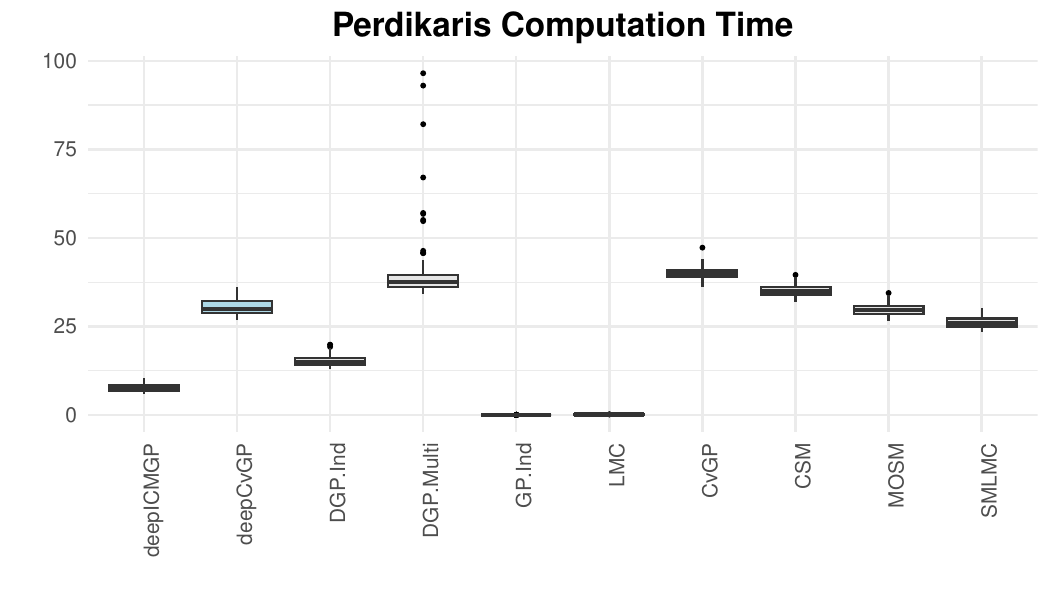}
  \end{subfigure}
  \vskip 0.1\baselineskip
  \begin{subfigure}[t]{0.49\linewidth}
    \centering
    \includegraphics[width=0.95\linewidth]{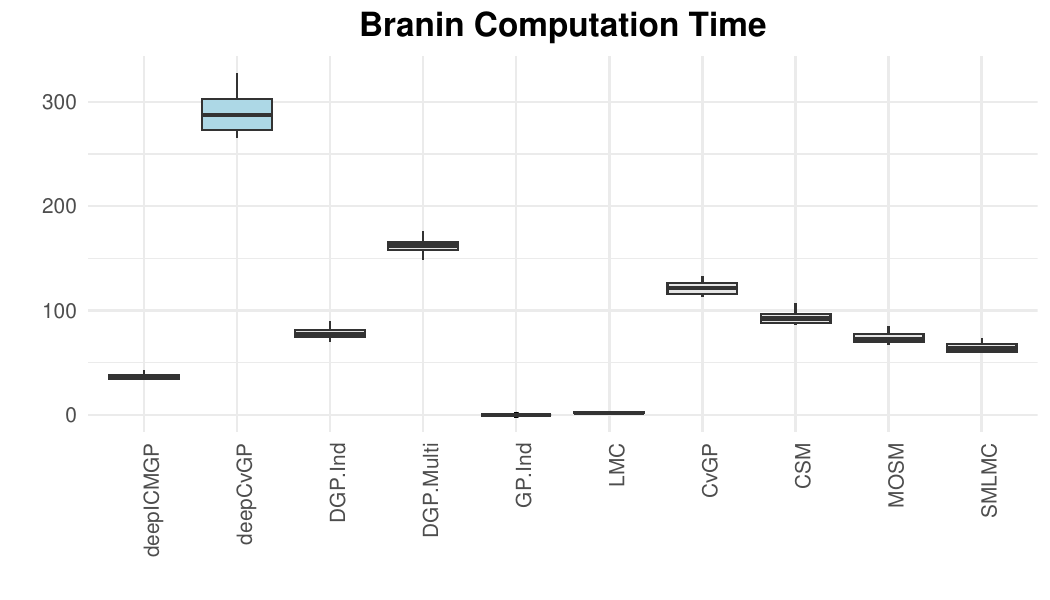}
  \end{subfigure}
  \hfill
  \begin{subfigure}[t]{0.49\linewidth}
    \centering
    \includegraphics[width=0.95\linewidth]{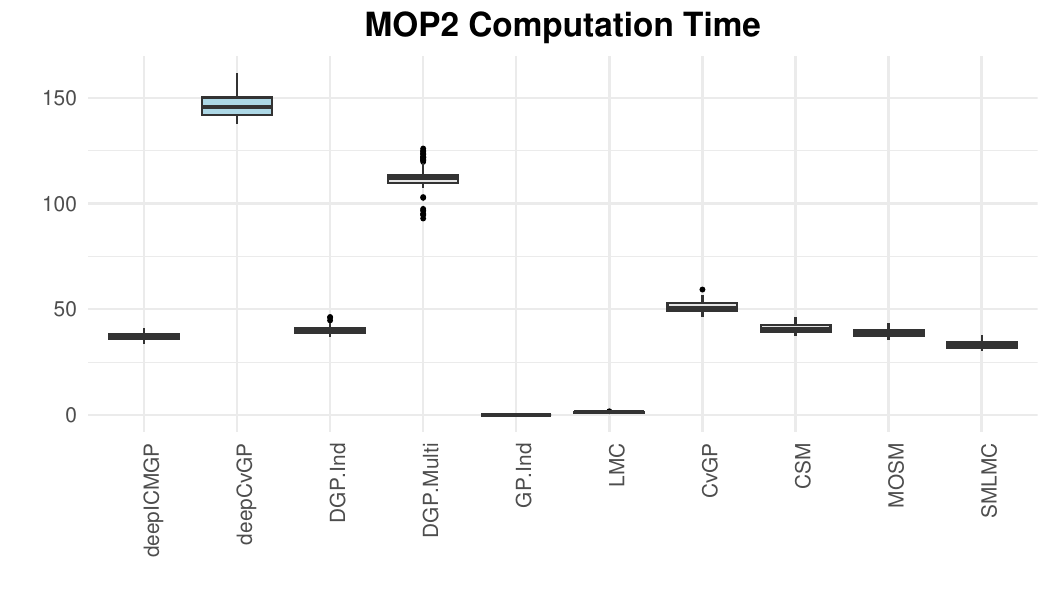}
  \end{subfigure}
  \vskip 0.1\baselineskip
  \begin{subfigure}[t]{0.49\linewidth}
    \centering
    \includegraphics[width=0.95\linewidth]{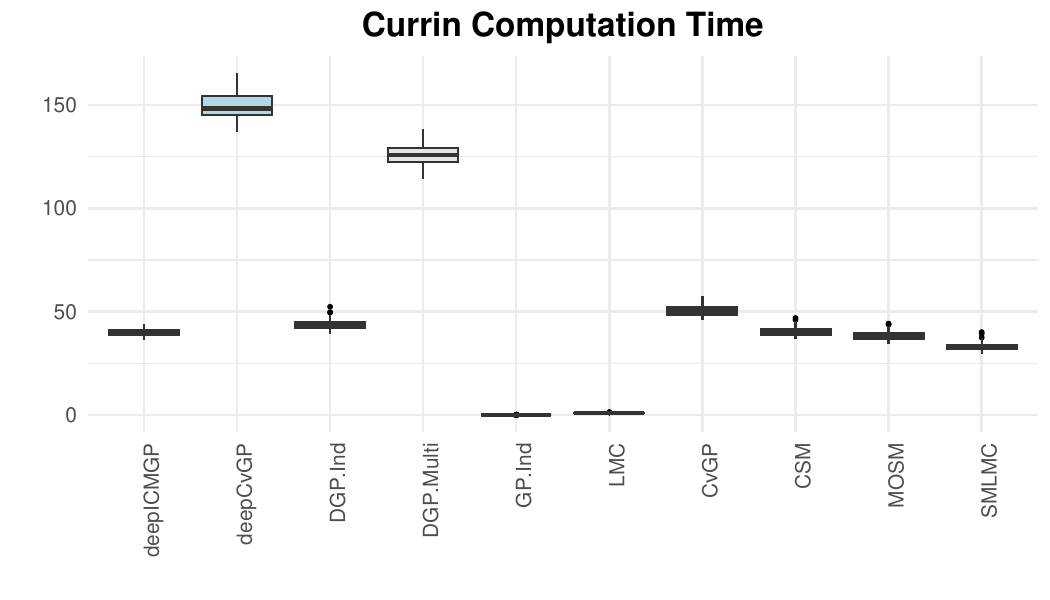}
  \end{subfigure}
  \hfill
  \begin{subfigure}[t]{0.49\linewidth}
    \centering
    \includegraphics[width=0.95\linewidth]{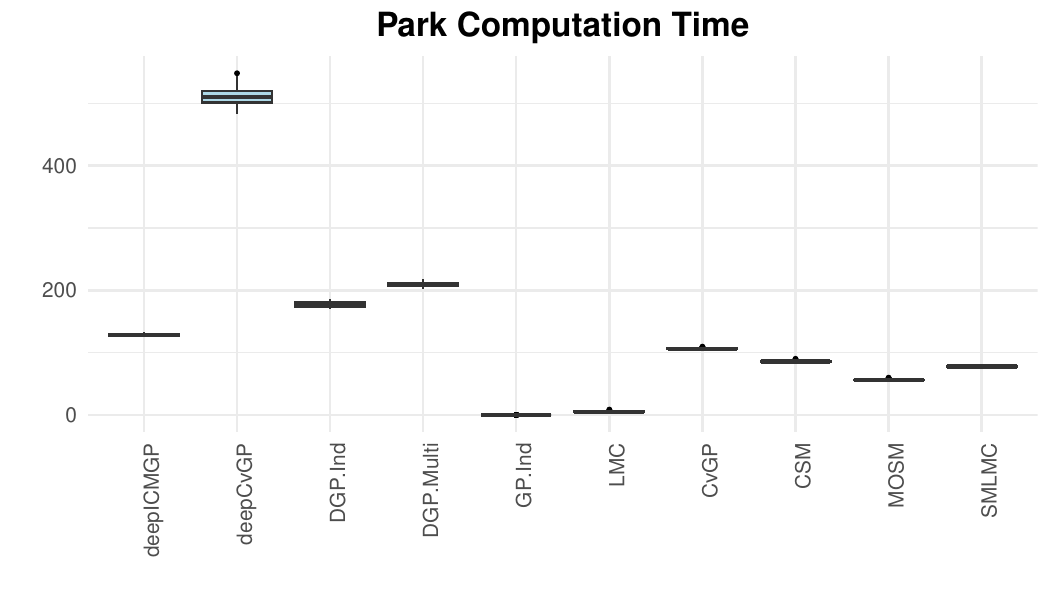}
  \end{subfigure}
  
  \caption{Computation time comparison of \texttt{deepICMGP} and competing methods across all examples. Boxplots summarize results from 100 repetitions with different training samples. }
  \label{fig:time_res}
\end{figure}

Figures \ref{fig:RMSE_res}--\ref{fig:time_res} present the boxplots of RMSE, CRPS, multivariate Score, and computational cost, respectively. Each result is based on 100 independent repetitions of the simulation, using different maximin LHDs for the training data. 

Overall, no single method dominates across all examples. From the RMSE (Figure \ref{fig:RMSE_res}) and CRPS (Figure \ref{fig:CRPS_res}) results, which assess individual outputs, \texttt{deepICMGP} generally performs well across all outputs and examples, frequently ranking in the top five (e.g., Forrester, Branin, Currin, and Park functions). In contrast, some methods, such as \texttt{deepCvGP}, \texttt{CSM}, \texttt{MOSM}, and \texttt{SMLMC}, show unstable performance. In certain cases, \texttt{deepICMGP} excels in one output but is less competitive in another (e.g., Perdikaris). From the multivariate scores (Figure \ref{fig:score_res}), which summarize the performance of all outputs, \texttt{deepICMGP} is highly competitive, often ranking in the top three, and frequently outperforming its DGP-based alternatives (\texttt{DGP.ind} and \texttt{DGP.multi}). In contrast, our alternative method \texttt{deepCvGP} performs less well, likely due to over-flexibility and a large number of parameters, leading to instability. The computational cost results (Figure \ref{fig:time_res}) show that \texttt{deepICMGP} is also computationally efficient, often ranking in the top three and, in most cases, faster than its DGP alternatives. As discussed in Section \ref{sec:posterior}, this efficiency comes from assuming a single lengthscale parameter per layer and marginalizing the coregionalization matrices, rather than estimating separate lengthscale parameters per node as in DGP.

In summary, \texttt{deepICMGP} offers strong predictive accuracy, reliable uncertainty quantification, and competitive computational efficiency, striking a good balance among these criteria.

\subsection{Simulation Study for Active Learning}\label{sec:num_alc}

In this section, we evaluate the performance of \texttt{deepICMGP} in a sequential design setting using the ALC criterion introduced in Section~\ref{sec:AL}, with the Branin function as a benchmark. The model is first trained on an initial design of size $n_0=30$ generated via maximin LHD, followed by the sequential acquisition of 10 additional points (final sample size $n=40$) using the ALC criterion. A set of 400 candidate and reference points, generated from a $20\times20$ grid, is used to select the optimal locations and compute the integrals.

For comparison, we also consider the independent deep GP model (\texttt{DGP.Ind}), where the ALC criterion \eqref{eq:alcMulti} is applied to $Q$ independent deep GPs, implying zero correlation across outputs.

\begin{figure}[htbp]
    \centering
    \includegraphics[width=\linewidth]{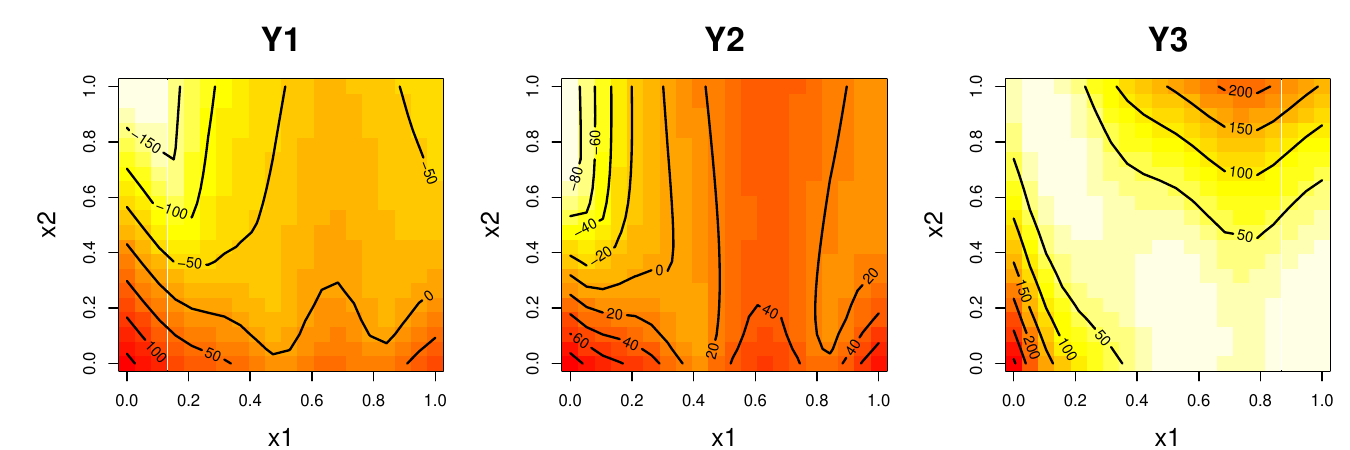}
    \caption{The true values of the synthetic Branin function.}
    \label{fig:ALC_mean}
\end{figure}

\begin{figure}[htbp]
    \centering
    \includegraphics[width=\linewidth]{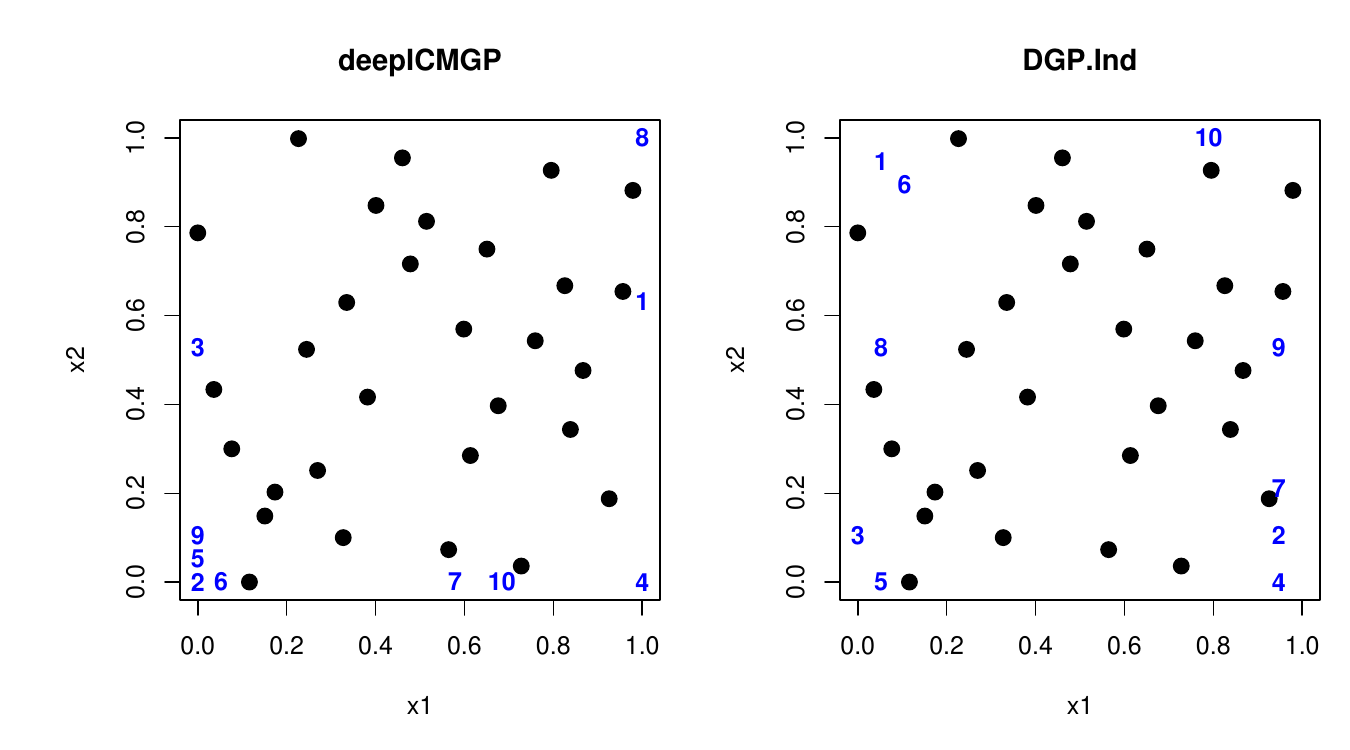}
    \caption{Active learning sample selection for \texttt{deepICMGP} (left) and \texttt{DGP.Ind} (right). Black dots represent the initial design of size 30, and blue numbers indicate the order in which additional points are sequentially selected.}
    \label{fig:ALC_plot1}
\end{figure}

Figure \ref{fig:ALC_mean} shows the true Branin function, and Figure \ref{fig:ALC_plot1} illustrates the sequentially selected design points under both methods. The ALC criterion of \texttt{deepICMGP} tends to concentrate samples in more wiggly regions, particularly for $y_1$ and $y_2$, which are of greater modeling interest. The joint modeling of \texttt{deepICMGP} enables improved prediction of $y_3$ from $y_1$ and $y_2$. By contrast, the ALC criterion of \texttt{DGP.Ind}, assuming independence, favors locations with overall high uncertainty across outputs, leading to a more space-filling design.

\begin{figure}[htbp]
  \centering
  \begin{subfigure}[t]{0.6\linewidth}
    \centering
    \includegraphics[width=\linewidth]{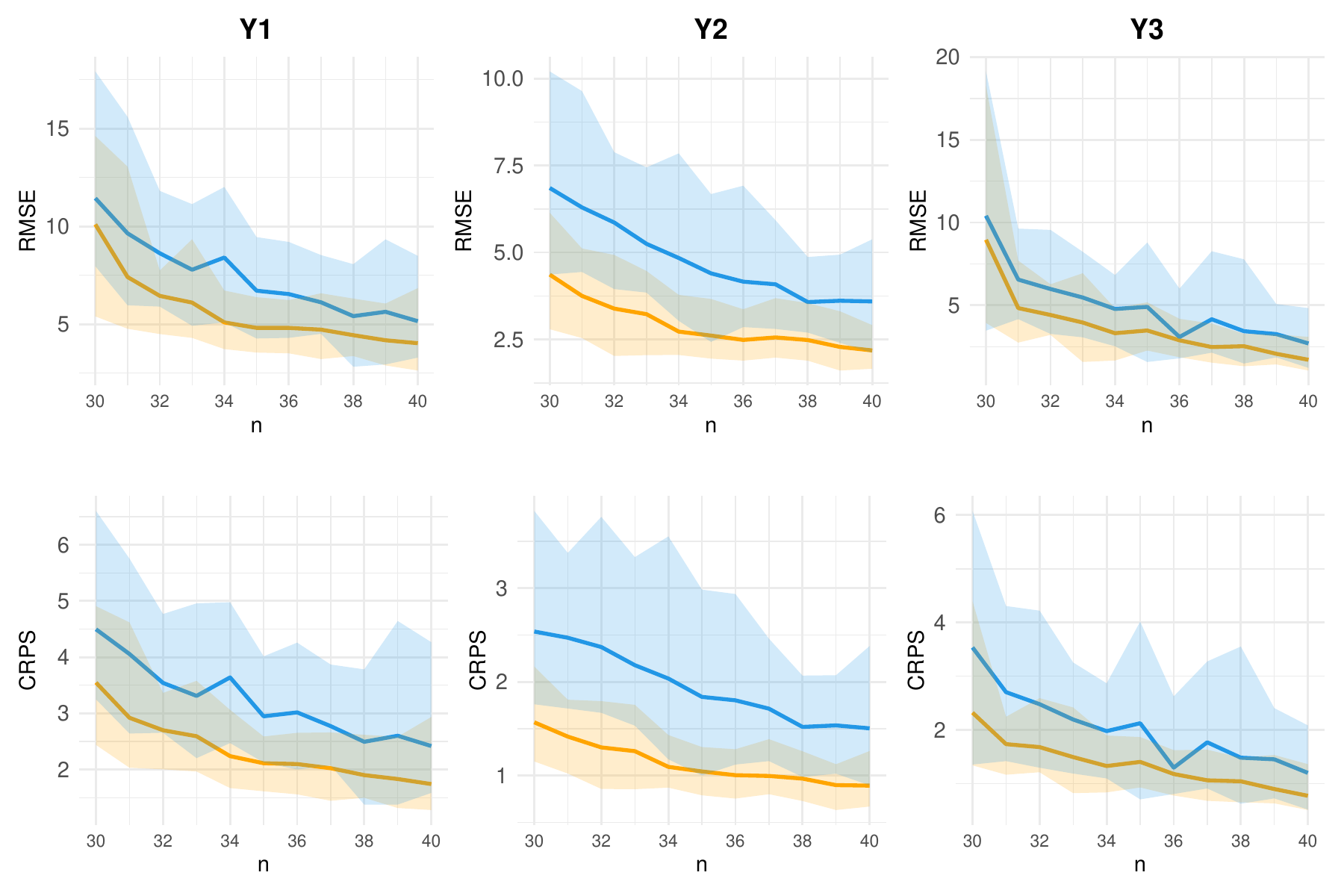}
  \end{subfigure}
  \hfill
  \begin{subfigure}[t]{0.39\linewidth}
    \centering
    \includegraphics[width=\linewidth]{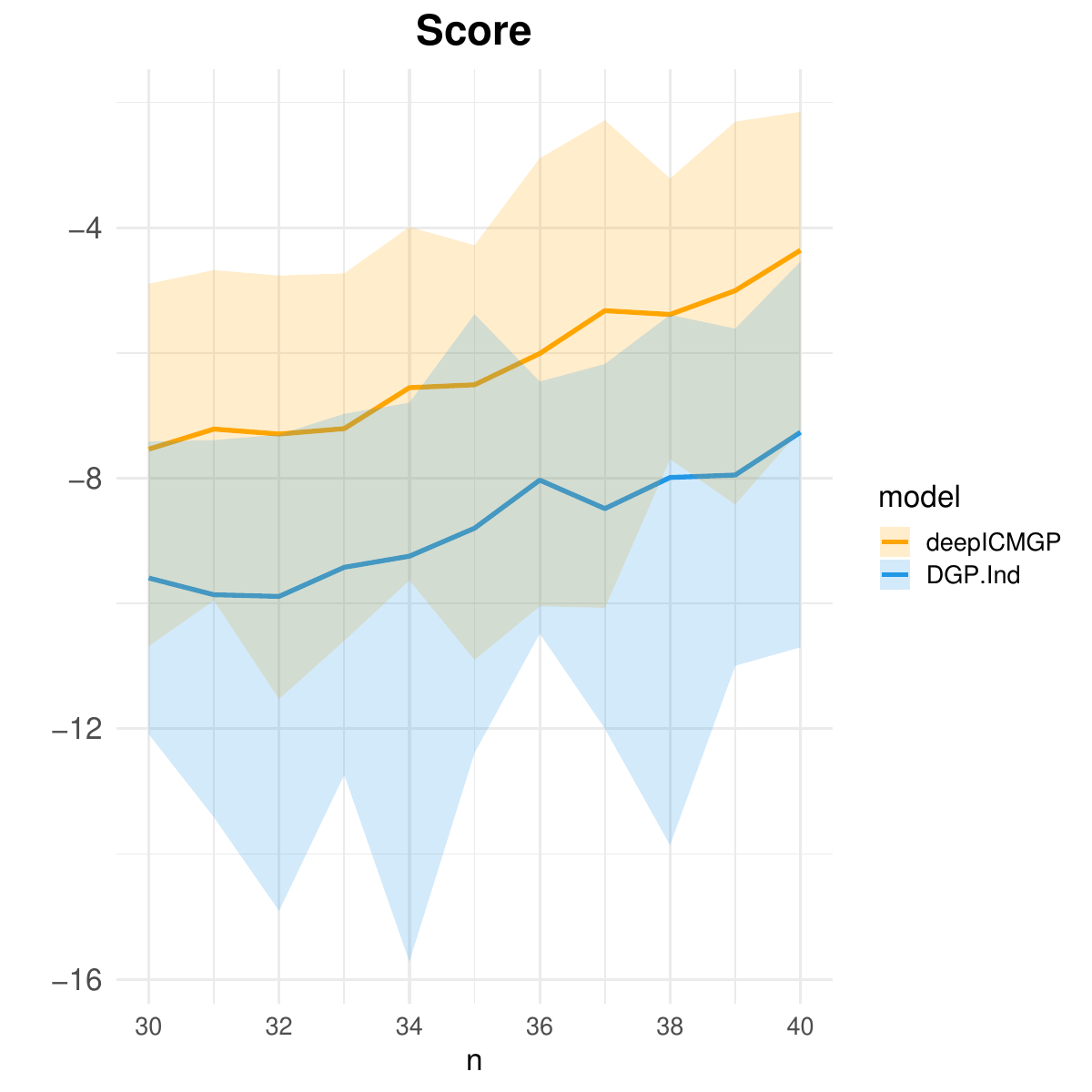}
  \end{subfigure}
  
  \caption{Active learning performance on the synthetic Branin function. Solid lines indicate mean values, and shaded regions represent 95\% confidence intervals. Left: RMSE (top) and CRPS (bottom). Right: Multivariate Score.}
  \label{fig:Branin_alc}
\end{figure}


We repeat the simulation 20 times with different initial designs. The results are summarized in Figure \ref{fig:Branin_alc}, which shows that \texttt{deepICMGP} consistently outperforms \texttt{DGP.Ind} across varying sample sizes. In particular, \texttt{deepICMGP} achieves lower RMSE and CRPS values while simultaneously obtaining higher multivariate scores, demonstrating both improved predictive accuracy and better overall uncertainty quantification.



\section{Case Study: Jet Engine Turbine Blade Simulation}\label{sec:real_data_analysis}
We apply the proposed emulator to a real-world case study of thermal stress analysis in a jet engine turbine blade. The turbine, a critical component of a jet engine, is typically made from nickel-based alloys. Evaluating its structural performance is essential, as excessive stress or deformation can lead to mechanical failure or blade–casing contact. For further details, see \citet{CARTER2005237}, \citet{sung2024}, and \citet{Heo02012025}.

\begin{figure}[t]
   \centering
   \begin{subfigure}[t]{0.32\linewidth}
     \centering
     \includegraphics[width=\linewidth]{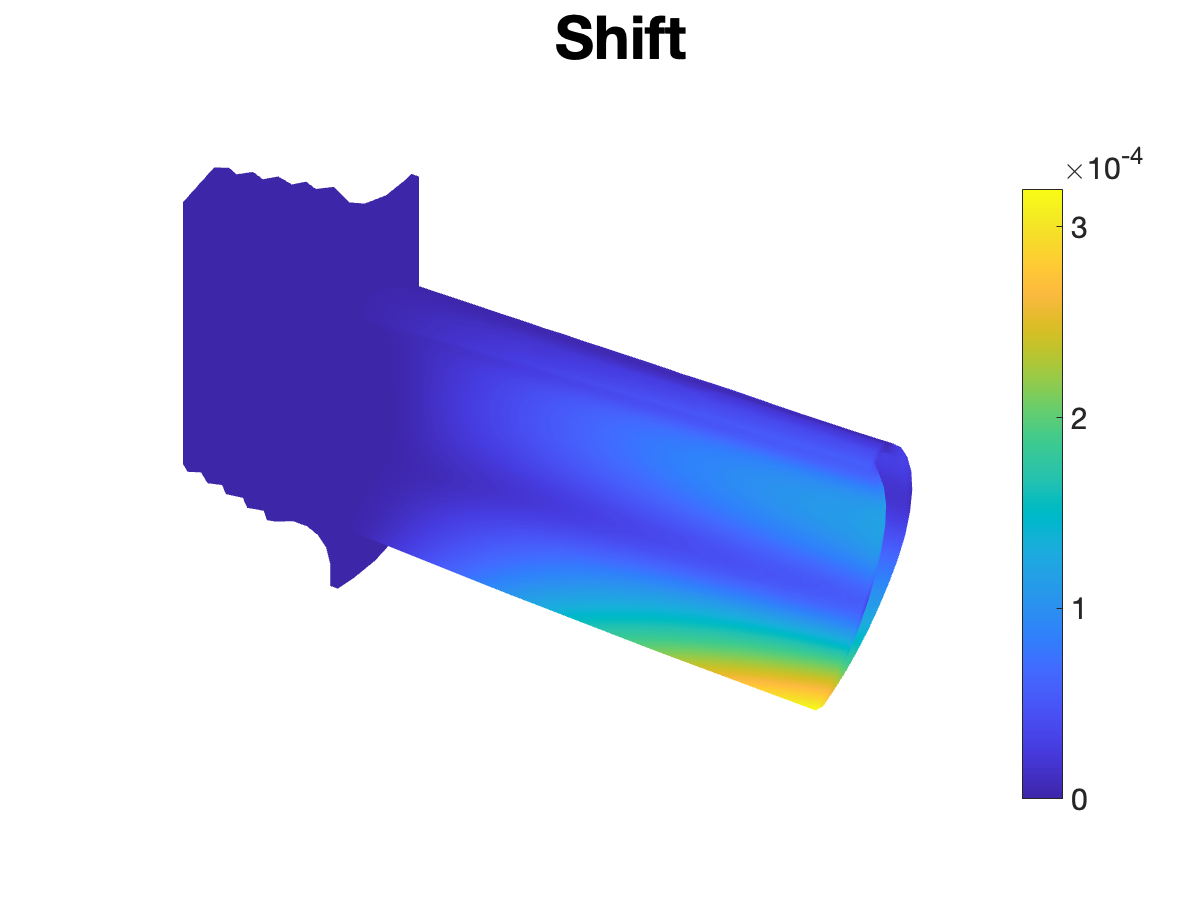}
   \end{subfigure}
   \begin{subfigure}[t]{0.32\linewidth}
     \centering
     \includegraphics[width=\linewidth]{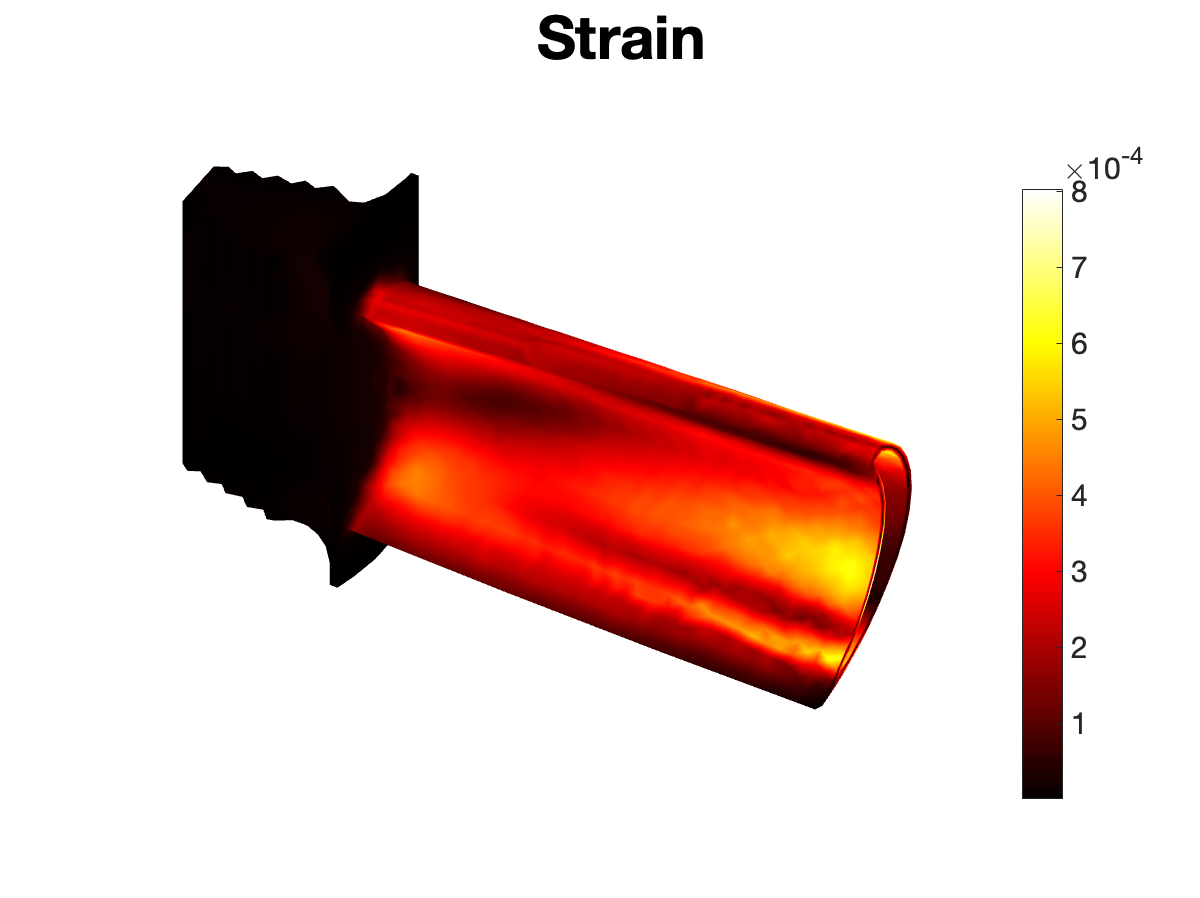}
   \end{subfigure}
   \begin{subfigure}[t]{0.32\linewidth}
     \centering
     \includegraphics[width=\linewidth]{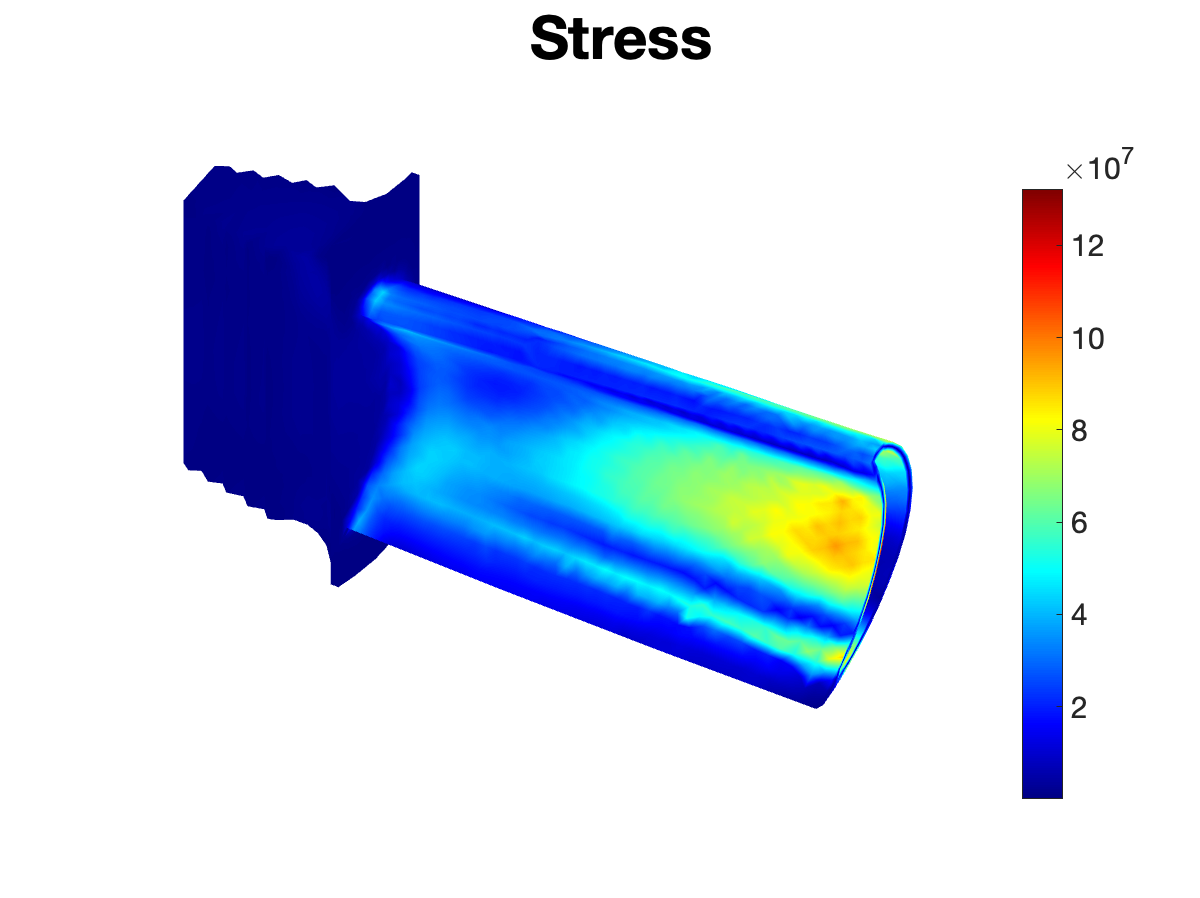}
   \end{subfigure}
  
   \caption{Illustration of the three outputs from the FEM simulation of the turbine blade at input location $\mathbf{x} = (0.5, 0.5)$, namely Shift (left), Strain (center), and Stress (right).}
   \label{fig:blade}
 \end{figure}

The problem is modeled as a static structural analysis and solved numerically using the finite element method (FEM) \citep{brenner2007fem}, implemented via the \texttt{Partial Differential Equation Toolbox} in \textsf{MATLAB} \citep{pde}. The model has two input variables: pressure loads applied to the pressure and suction sides of the blade, sharing the same input space $\mathbf{x} \in [0.25, 0.75]^2$, corresponding to a range of $0.25$–$0.75$ MPa. We focus on three scalar outputs: structural displacement (\textit{Shift}), stress induced by thermal loads (\textit{Stress}), and strain representing relative deformation (\textit{Strain}). Figure \ref{fig:blade} demonstrates the turbine blade simulation for the three outputs, with the quantities of interest defined as the maximum values at the blade tip.

Following the setup in Section \ref{sec:numsetup}, we generate $20$ training inputs using a maximin LHD and $400$ test input locations on a uniform grid over $\mathcal{X} \in [0.25,0.75]^2$. This experiment is repeated $30$ times with different random training designs. For active learning evaluation, we use $n_0=20$ initial design points, followed by sequential acquisition of $10$ additional points via the ALC criterion introduced in Section \ref{sec:AL}. As in Section \ref{sec:num_alc}, the experiment is repeated $20$ times with different initial designs. We compare our approach with the ALC criterion applied to \texttt{DGP.Ind}.

\begin{figure}[htbp]
  \centering
  \begin{subfigure}[t]{0.39\linewidth}
    \centering
    \includegraphics[width=\linewidth]{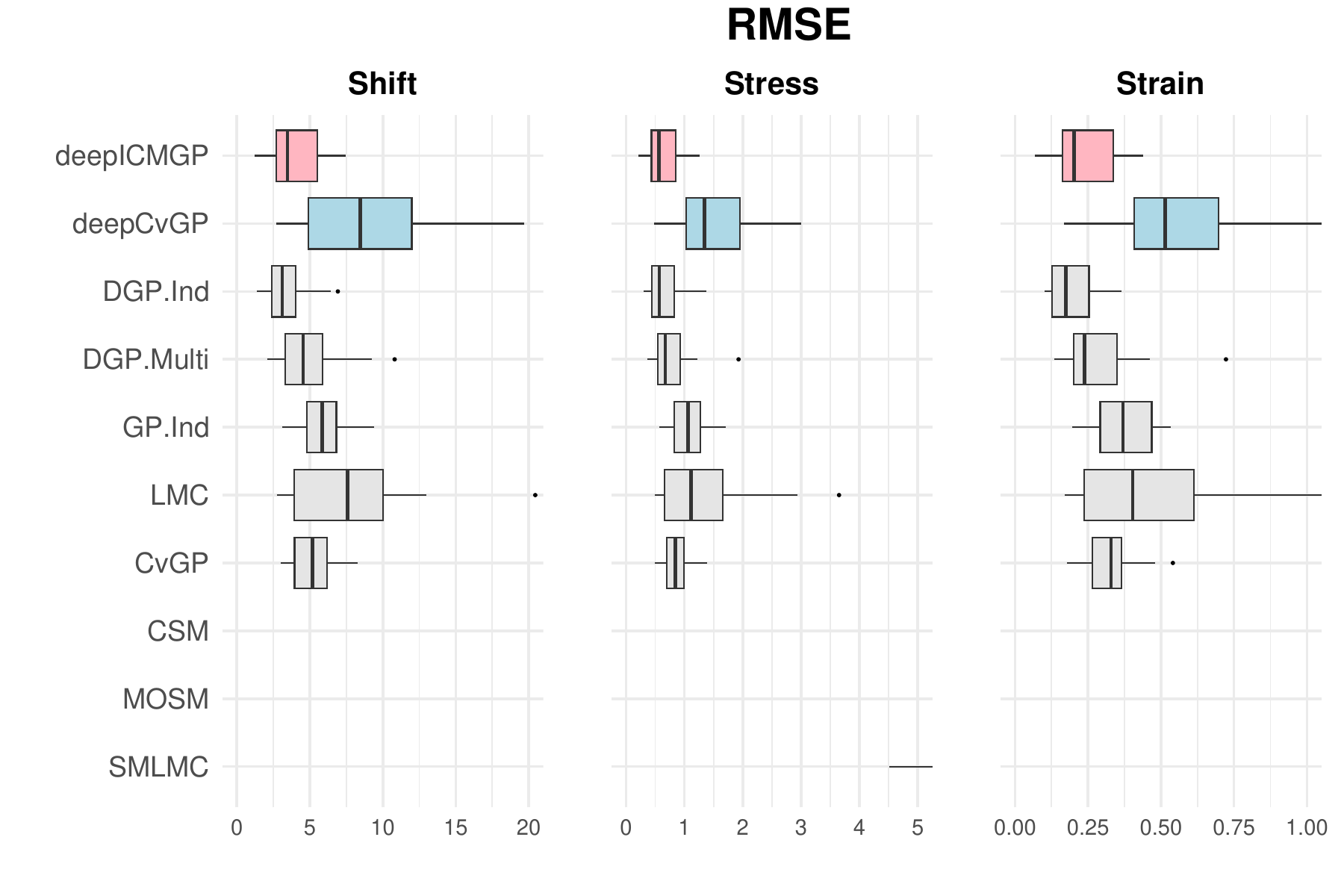}
  \end{subfigure}
  \hfill
  \begin{subfigure}[t]{0.39\linewidth}
    \centering
    \includegraphics[width=\linewidth]{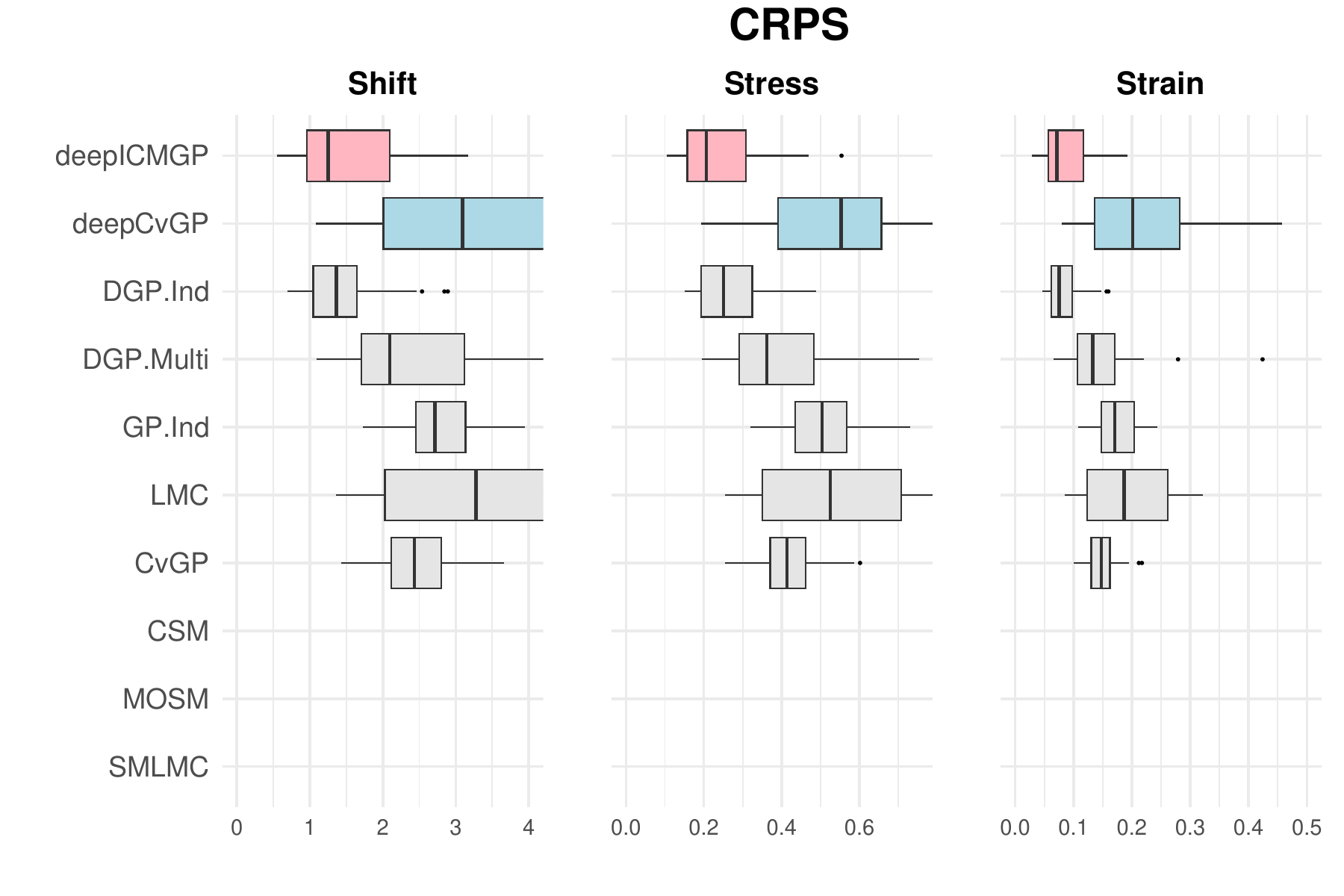}
  \end{subfigure}
  \hfill
  \begin{subfigure}[t]{0.2\linewidth}
    \centering
    \includegraphics[width=\linewidth]{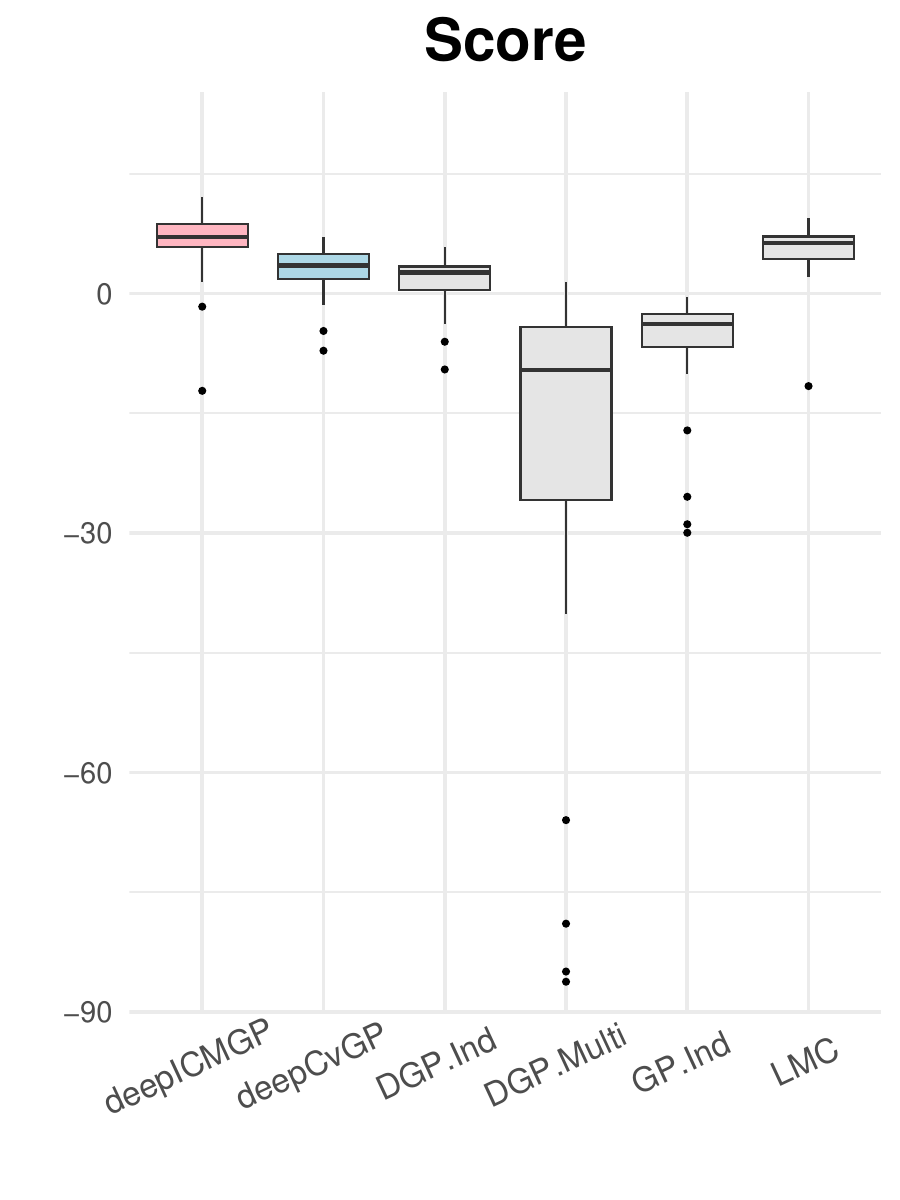}
  \end{subfigure}
  
  \caption{Comparison of \texttt{deepICMGP} and competing methods on the turbine blade simulation. Boxplots are based on 30 repetitions, showing RMSE (left), CRPS (center), and Multivariate Score (right). Some extreme boxplots are omitted as their values fall outside the displayed range.}
  \label{fig:blade_res}
\end{figure}

Prediction results are shown in Figure \ref{fig:blade_res}. \texttt{DeepICMGP} outperforms all other methods across all metrics, achieving results comparable to \texttt{DGP.Ind} in RMSE and CRPS. While \texttt{DGP.Ind} performs competitively for individual outputs, its multivariate score is notably lower. In contrast, \texttt{deepICMGP} achieves the highest multivariate Score, comparable to \texttt{LMC}, indicating superior modeling of joint uncertainty and dependencies among outputs.

\begin{figure}[htbp]
  \centering
  \begin{subfigure}[t]{0.6\linewidth}
    \centering
    \includegraphics[width=\linewidth]{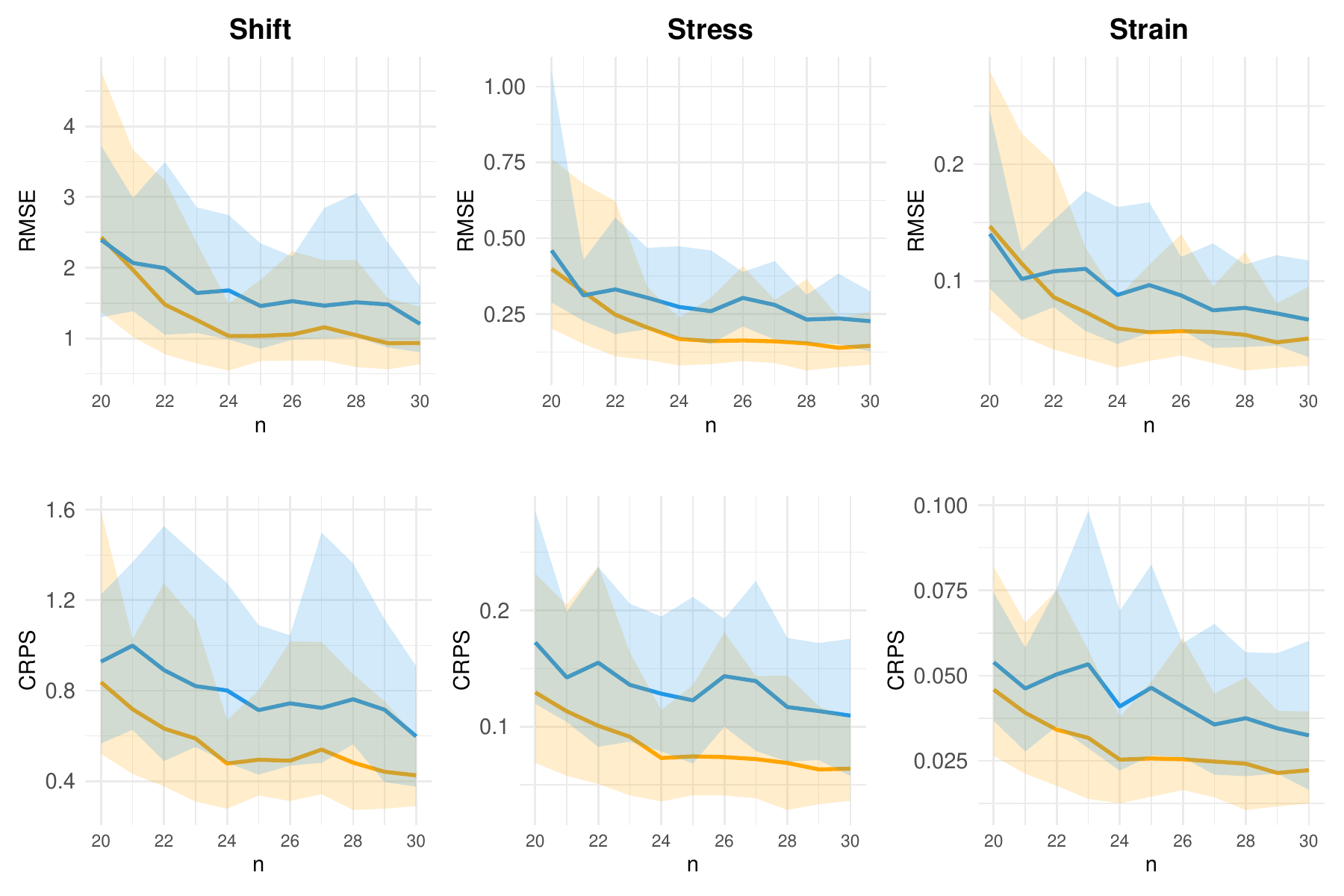}
  \end{subfigure}
  \hfill
  \begin{subfigure}[t]{0.39\linewidth}
    \centering
    \includegraphics[width=\linewidth]{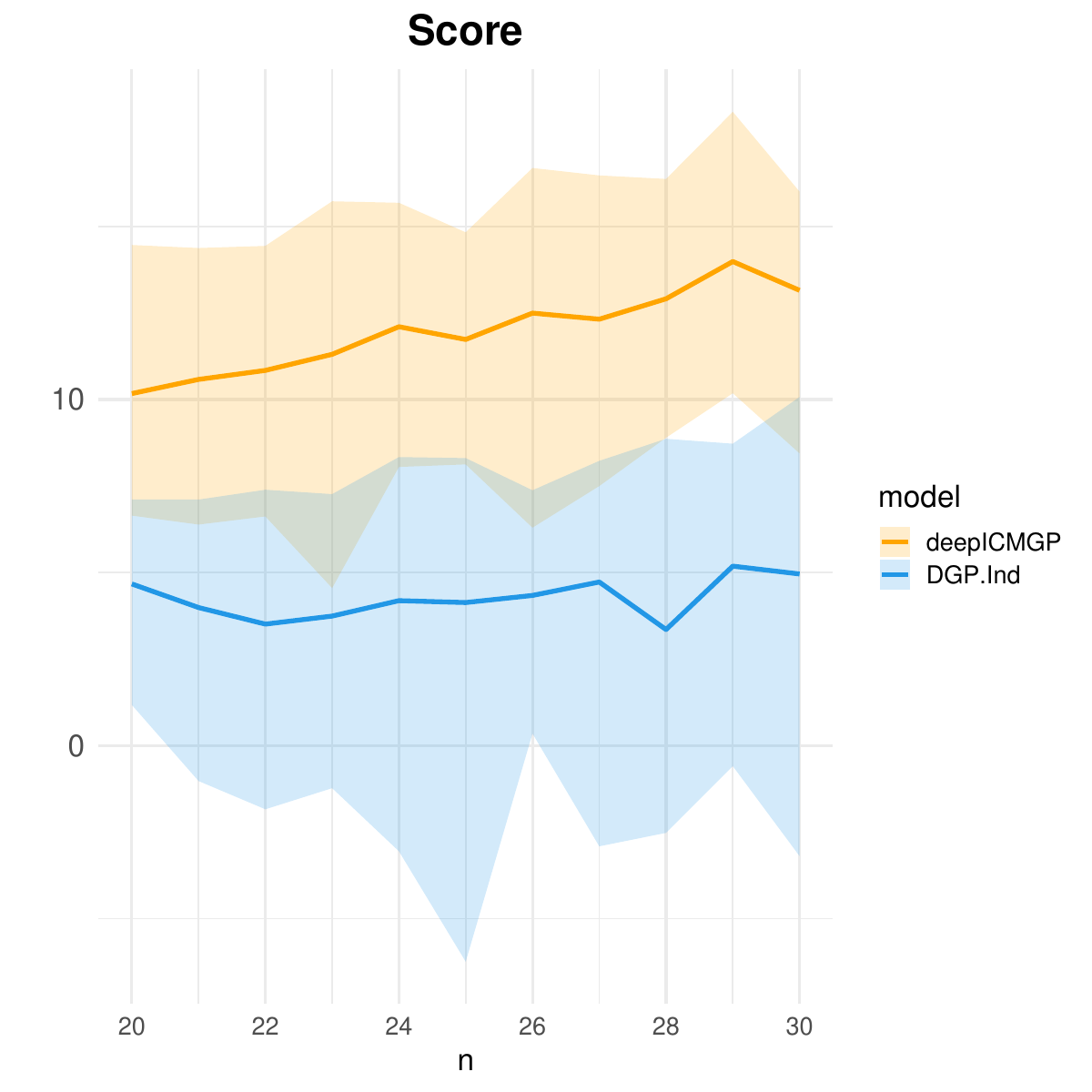}
  \end{subfigure}
  
  \caption{Active learning performance on the turbine blade simulation. Solid lines indicate mean values, and shaded regions represent 95\% confidence intervals. Left: RMSE (top) and CRPS (bottom). Right: Multivariate Score.}
  \label{fig:blade_alc_res}
\end{figure}

Active learning results are presented in Figure \ref{fig:blade_alc_res}. Both RMSE and CRPS from \texttt{deepICMGP} are consistently better than those from \texttt{DGP.Ind}. Although \texttt{deepICMGP} starts with slightly higher RMSE and CRPS for \textit{Strain} in the early iterations, performance improves steadily and surpasses \texttt{DGP.Ind} in later steps. Furthermore, the multivariate scores of \texttt{deepICMGP} remain consistently higher, highlighting its stability and robustness during sequential design. 


\section{Conclusion}\label{sec:conclusion}

Multi-output computer simulations are prevalent across many fields, particularly in engineering applications. In this paper, we proposed the deepICMGP model, a deep GP framework that incorporates intrinsic coregionalization to efficiently emulate multi-output computer simulations. By using a single lengthscale parameter per layer and marginalizing the coregionalization matrices, deepICMGP reduces computational complexity while retaining flexibility to model complex dependencies across outputs. In addition, we introduced an active learning criterion tailored for deepICMGP to identify informative input locations, thereby enhancing prediction accuracy and improving the efficiency of sampling for multi-output computer experiments. Our results demonstrate that deepICMGP achieves competitive or superior predictive accuracy compared with other multi-output GP and DGP alternatives.



Several directions are compelling for future work. Like many GP models, deepICMGP encounters challenges when applied to large-scale datasets. Existing methods such as Vecchia approximations \citep{vecchia2017Katzfuss, Sauer03072023Vecchia} and inducing point techniques \citep{SparseGP2005Snelson} may help address these issues. In addition, future work could investigate co-active subspace methods \citep{AS2014Constantine, rumsey2025co} within the deepICMGP framework to mitigate the curse of dimensionality for high-dimensional data and enable joint analysis of two or more computer models, allowing for a thorough exploration of the alignment of their respective gradient spaces.

\vspace{0.5cm}
\noindent\textbf{Acknowledgements}
The authors gratefully acknowledge funding from NSF DMS 2338018.


\begin{appendices}

\section{Synthetic Functions}\label{sup: functions}
We provide all the explicit functions used in the numerical studies described in Section \ref{sec:numerical}.
\begin{itemize}
    \item Forrester Function \citep{forrester2008}
    \begin{align*}
        f_1(x)&=(6x-2)^2\sin(12x-4)\\
        f_2(x)&=0.5f_1(x)+10(x-0.5)+5,\,\forall x \in [0,1]
    \end{align*}
    \item Convolved Squared Exponential Kernel (\cite{negative2022})
    \begin{align*}
        f_1(x)&=5\sin\left(\frac{3x}{2}\right)\\
        f_2(x)&=5\sin(x)-3\\
        f_3(x)&=\frac{x^2}{10}-5,\,\forall x \in [0,10]
    \end{align*}
    \item Damped Wave Function
    \begin{align*}
        f_1(x)&=5e^{-10x}\left(\cos(10\pi x-1)+\sin(10\pi x-1)\right)-0.2\\
        f_2(x)&=6e^{-5x}\left(\cos(10\pi x-1)+\sin(5\pi x-1)\right)-0.1\\
        f_3(x)&=4e^{-15x}\left(\cos(5\pi x-1)+\sin(15\pi x-1)\right)-0.3, \,\forall x \in [0,1]
        \end{align*}
    \item Perdikaris Function \citep{perdikaris2017}
    \begin{align*}
        f_1(x)&=\sin(8\pi x)\\
        f_2(x)&=(x-\sqrt{2})f_1(x)^2, \,\forall x \in [0,1]
    \end{align*}
    \item Branin Function \citep{forrester2008}
    \begin{align*}
        f_1(\mathbf{x})&=f_2(1.2(\mathbf{x}+2))-3x_2+1\\
        f_2(\mathbf{x})&=10\sqrt{f_3(x)}+2(x_1-0.5)-3(3x_2-1)-1\\
        f_3(\mathbf{x})&=\left(\frac{-1.275x_1^2}{\pi^2}+\frac{5x_1}{\pi}+x_2-6\right)^2+\left(10-\frac{5}{4\pi}\right)\cos(x_1)+10, \, \mathbf{x} \in [-5,10]\times [0,15]
    \end{align*}
    \item The MOP2 Problem \citep{Fonseca1995MultiobjectiveGA,SVENSON2016250}
    \begin{align*}
    f_1(\mathbf{\mathbf{x}})&=1-exp\left[-\sum_{i=1}^2(x_i-\frac{1}{\sqrt{2}})^2\right]\\
    f_2(\mathbf{\mathbf{x}})&=1-exp\left[-\sum_{i=1}^2(x_i+\frac{1}{\sqrt{2}})^2\right], \,\mathbf{x} \in [-2,2]^2
    \end{align*}
    \item Currin Function \citep{currin1988, xiong2012}
    \begin{align*}
        f_1(\mathbf{x})&=\left(1-\exp\left(-\frac{1}{2x_2}\right)\right)\frac{2300x_1^3+1900x_1^2+2092x_1+60}{100x_1^3+500x_1^2+4x_1+20}\\
        f_2(\mathbf{x})&=\frac{1}{4}\left[f_1\left(x_1+\frac{1}{20},x_2+\frac{1}{20}\right)
        +f_1\left(x_1+\frac{1}{20}, \max \left(0,x_2-\frac{1}{20}\right)\right)\right]\\
        &+\frac{1}{4}\left[f_1\left(x_1-\frac{1}{20},x_2+\frac{1}{20}\right)+f_1\left(x_1-\frac{1}{20}, \max \left(0,x_2-\frac{1}{20}\right)\right)\right], \,\mathbf{x} \in [0,1]^2
    \end{align*}

    \item Park Function \citep{park1991tuning,xiong2012}
    \begin{align*}
        f_1(\mathbf{x})&=\frac{x_1}{2}\left(\sqrt{\frac{1+(x_2+x_3^2)x_4}{x_1^2}}-1\right)+\left(x_1+3x_4\right)\exp(1+\sin(x_3))\\
        f_2(\mathbf{x})&=\left(\frac{1+\sin(x_1)}{10}\right)f_1(\mathbf{x})-2x_1 + x_2^2 + x_3^2 + 0.5, \,\mathbf{x} \in [0,1]^4
    \end{align*}
    
\end{itemize}
\end{appendices}

\bibliography{ref}

\end{document}